\documentclass[lettersize,journal]{IEEEtran}
\usepackage{amsmath,amsfonts}
\usepackage{algorithmic}
\usepackage{algorithm}
\usepackage{array}
\usepackage[caption=false,font=normalsize,labelfont=sf,textfont=sf]{subfig}
\usepackage{textcomp}
\usepackage{stfloats}
\usepackage{float} 
\usepackage{url}
\usepackage{verbatim}
\usepackage{graphicx}
\usepackage{cite}
\usepackage{arydshln}
\usepackage{multirow}
\usepackage{booktabs}
\usepackage{makecell}
\usepackage{amssymb}
\usepackage{subfig}
\usepackage{wrapfig}
\usepackage[colorlinks=true, linkcolor=blue, citecolor=blue, urlcolor=blue]{hyperref}
\newcommand{\figref}[1]{Fig.~\ref{#1}}
\newcommand{\tabref}[1]{Table~\ref{#1}}
\begin{document}

\title{Act in Collusion: Distributed Multi-Target Backdoor Attacks in Federated Learning}
\author{Tao Liu, Dapeng Man, Jiguang Lv, Chen Xu, Weiye Xi, Huanran Wang, Yuhang Zhang, Tianming Zhao,Wu Yang
\thanks{Manuscript received ----------------; revised ----------------; accepted ----------------. Date of publication ----------------; date of current version ----------------. This work was supported in part by the National Natural Science Foundation of China (No.62272127, No.62406086, No.62572144), the Joint Funds of the National Natural Science Foundation of China (No.U22A2036, No.U21B2019), and Basic Research Projects of the Central Universities and Colleges (3072025ZN0602).\textit{(Corresponding author: Jiguang Lv.)}}
\thanks{The authors are with the College of Computer Science and Technology, Harbin Engineering University, Harbin 150001, China (e-mail: ltaoheu@163.com; mandapeng@hrbeu.edu.cn; lvjiguang@hrbeu.edu.cn; chen.xu@hrbeu.edu.cn; xiweiye@hrbeu.edu.cn; huanran.wang@hrbeu.edu.cn; z1459384884@hrbeu.edu.cn; yangwu@hrbeu.edu.cn).}
\thanks{Copyright (c) 2026 IEEE. Personal use of this material is permitted. However, permission to use this material for any other purposes must be obtained from the IEEE by sending a request to pubs-permissions@ieee.org.}
\thanks{Digital Object Identifier --.----/JIOT.----.----}}

\markboth{Journal of \LaTeX\ Class Files,~Vol.~14, No.~8, August~2021}%
{Shell \MakeLowercase{\textit{et al.}}: A Sample Article Using IEEEtran.cls for IEEE Journals}

\maketitle

\begin{abstract}
Federated learning (FL) is widely used in Internet-of-Things (IoT) systems, but its distributed training process also exposes it to backdoor attacks. Existing studies mainly consider single-target or centralized multi-target settings, while coordinated distributed multi-target attacks remain underexplored. In practical IoT scenarios, one adversarial entity may control multiple distributed malicious clients and assign each client distinct triggers and  target labels. Under this setting, existing distributed backdoor methods often fail to preserve the effectiveness of all backdoors because malicious updates conflict during aggregation. To address this issue, we propose a Distributed Multi-Target Backdoor Attack (DMBA) for FL. DMBA introduces a Backdoor Replay (BR) mechanism to reduce discrepancies among malicious gradients and a Channel-Frequency Composite Trigger (CFCT) strategy to improve trigger distinguishability and alleviate local interference. Experiments on multiple datasets show that DMBA ensures attack success rates above 80\% for all implanted backdoors, whereas some baseline backdoors fall below 50\% and may even approach 0.

\end{abstract}

\begin{IEEEkeywords}
Federated learning (FL), Internet-of-Things (IoT) systems, distributed multi-target backdoor, channel-frequency composite trigger, backdoor replay.
\end{IEEEkeywords}

\section{Introduction}
\IEEEPARstart{I}{n} recent years, Federated Learning (FL) has been increasingly applied to various Internet of Things (IoT)-based intelligent systems, such as healthcare~\cite{sheller2020federated}, autonomous driving~\cite{chellapandi2023survey}, and industrial automation~\cite{imteaj2021survey}. As IoT devices collect massive amounts of data locally, FL has emerged as a viable distributed learning paradigm~\cite{konevcny2016federated}, enabling decentralized devices to collaboratively train a global model without the need to share raw data. A central server aggregates model updates from client nodes~\cite{mcmahan2017communication}. This paradigm not only enhances user data privacy~\cite{wang2023blockchain} but also reduces data transmission costs~\cite{yang2019federated}, effectively boosting the performance of deep learning models in IoT systems.

However, the distributed nature of FL introduces significant security challenges~\cite{zhang2022neurotoxin}, especially backdoor attacks~\cite{wang2020attack}. In a backdoor attack, the adversary implants a trigger into the model so that the compromised model behaves normally on clean inputs but produces attacker-desired outputs when the trigger is present~\cite{baruch2019little}. This threat is particularly concerning in IoT-based FL systems, such as intelligent transportation~\cite{lu2024towards}, smart surveillance with distributed cameras~\cite{ryan2025smart}, and industrial inspection~\cite{zhukabayeva2025cybersecurity}, where large numbers of edge devices participate in training and device compromise may occur in practice.

Existing studies on backdoor attacks mainly focus on classification tasks~\cite{gu2017badnets} and can generally be divided into single-target attacks~\cite{shafahi2018poison},\cite{salem2022dynamic},\cite{chen2017targeted} and multi-target attacks~\cite{kwon2022multi},\cite{xue2022imperceptible},\cite{hou2022m}. Single-target attacks are easier to deploy and usually more effective, but their assumptions are limited. Multi-target attacks offer broader attack coverage, yet most of them are studied in centralized settings. A common limitation of these studies is that they do not adequately consider a practical FL scenario in which multiple distinct backdoor objectives are injected through different distributed clients.

In practical IoT environments~\cite{feng2022fiba}, a more realistic setting is that a single adversarial entity compromises and coordinates multiple malicious clients, and assigns each client a distinct trigger-target objective. This setting differs fundamentally from a conventional centralized multi-target attack. Although all malicious objectives are coordinated by one adversary, they still have to be executed through separate local training processes and uploaded as different client updates under the FL protocol. As a result, the attack becomes more stealthy at the client level, but also more difficult to optimize globally.
\begin{figure}[!t]
    \centering 
    \includegraphics[width=1\linewidth]{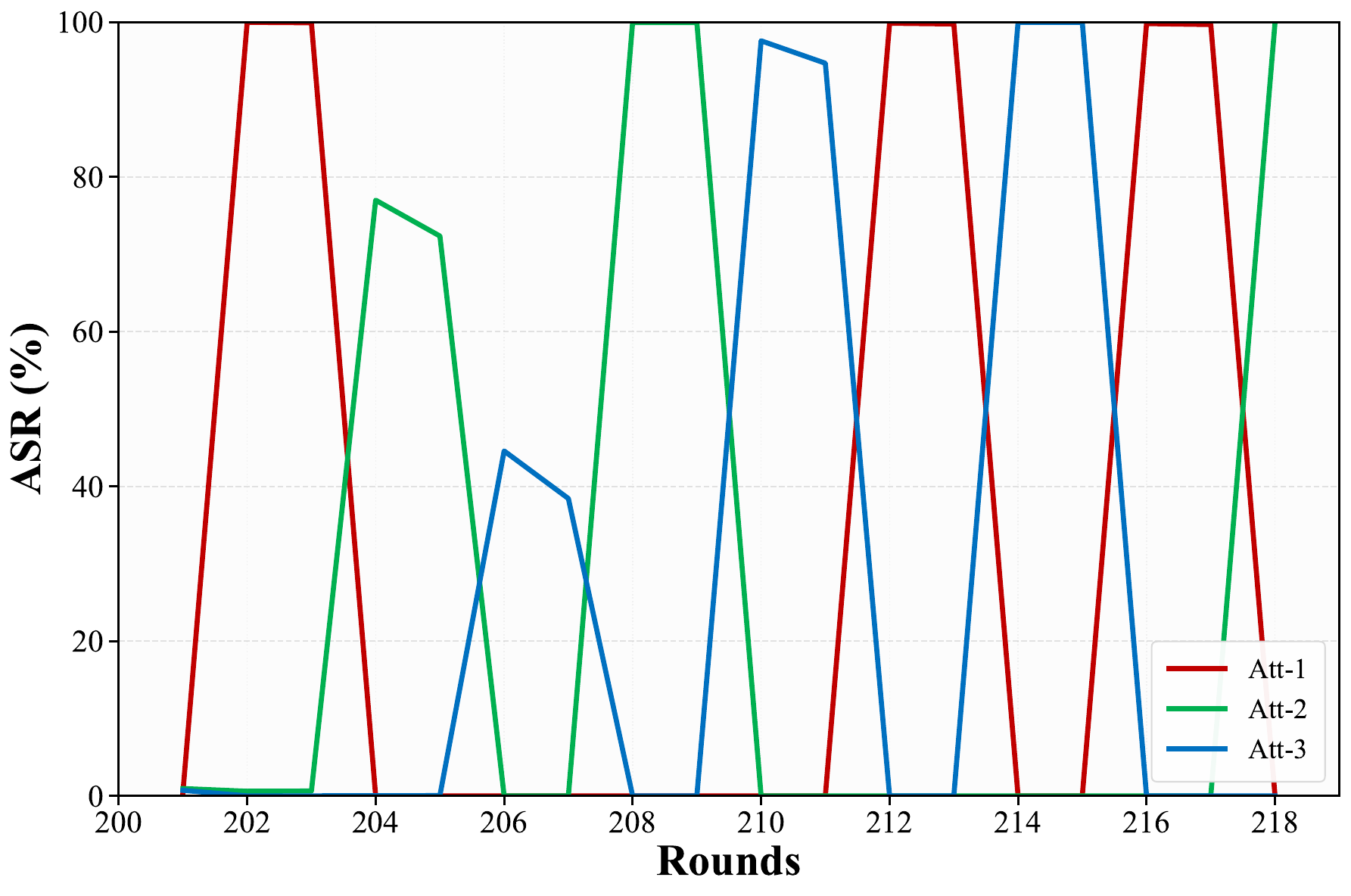} 
    \caption{Attack performance of the distributed backdoor attack on the threat model.} 
    \label{fig1}
\end{figure}
A key challenge in this setting is that malicious clients carrying different backdoor objectives may generate significantly different updates, which leads to gradient-direction conflicts during aggregationn~\cite{bagdasaryan2020backdoor}. To investigate this practical yet underexplored scenario, we define a distributed multi-target threat model and evaluate the performance of existing distributed backdoor attack methods~\cite{xie2019dba} under this setting. For a preliminary illustration, we consider a case with three malicious clients. As shown in \figref{fig1}, once a new backdoor is injected, previously implanted backdoors may weaken substantially or even fail.

To address this problem, we propose a Distributed Multi-Target Backdoor Attack (DMBA). DMBA mitigates gradient conflicts during FL training through two complementary designs. First, we introduce a Backdoor Replay (BR) mechanism to guide the local training of malicious clients and reduce the discrepancy among their uploaded malicious gradients. Second, we propose a Channel-Frequency Composite Trigger (CFCT) strategy to improve the distinguishability of different triggers and alleviate local interference among backdoor objectives.

Our contributions are summarized as follows:
\begin{itemize}
    
    \item We propose a practical distributed multi-target threat model and a novel backdoor attack, DMBA, for FL. This setting captures a realistic security risk in IoT-based FL systems and reveals a new vulnerability caused by coordinated distributed backdoor injection.
    \item We propose a backdoor replay mechanism and a channel-frequency composite trigger strategy. The BR mechanism stabilizes malicious training by reducing gradient conflicts during aggregation, while the CFCT strategy improves trigger distinguishability to alleviate conflicts during local training.
    \item Experiments on multiple datasets show that DMBA guarantees more than 80\% ASR for all backdoors, whereas some baseline backdoors fall below 50\% or even approach 0. The results also demonstrate strong trigger stealthiness and resistance to advanced defenses.
\end{itemize}

\section{Related work} 
\label{gen_inst}

Backdoor attacks, initially introduced by Gu et al.~\cite{gu2017badnets} as BadNets, involve injecting malicious triggers into neural networks during training to manipulate model behavior at test time. These attacks were quickly developed and expanded upon in the literature, with subsequent studies focusing on different strategies for increasing attack effectiveness and stealth. Recently, the field has evolved to address more complex attack scenarios, with a particular focus on multi-target backdoors and attacks in distributed computing environments such as FL.
\subsection{Backdoor attacks} 
\label{2.1}
\subsubsection{Multi-target backdoor attacks}

The objective of multi-target backdoor attacks is to implant several backdoors into a single model, each associated with a distinct trigger and target label. Early studies\cite{kwon2022multi}, Kwon et al.\cite{kwon2020multi} explored both fixed and position-variable multi-trigger strategies, introducing different backdoors in separate models, which, strictly speaking, is merely a multi-backdoor attack. Barni et al.\cite{barni2019new} initiated the first dual-target backdoor attack within the same model, using triangle and ramp signals as triggers for two target labels, and to enhance stealthiness, they adopted a clean-label setting. Building on this, Xue et al.\cite{xue2020one} extended the concept to multiple target labels, proposing the ONE-to-N attack, which activates multiple targets using a single pattern with varying transparency. Hou et al.\cite{hou2022m} further expanded this framework with the M-to-N attack, which enables multiple targets to be activated using different triggers. This approach is suitable for scenarios where the attacker has limited information and also employs semantic backdoors to improve stealth. More recently, Xue et al.\cite{xue2022imperceptible} shifted the trigger to the frequency domain, utilizing Discrete Cosine Transform (DCT) and steganography to realize imperceptible N-to-N attacks. This method embeds tiny triggers in different color channels, thus activating multiple targets without being detected by human observers.

Despite recent progress, these attacks assume centralized control over all data, which doesn't hold in FL. In FL, attackers can only poison local data or models, often leading to ineffective backdoors after aggregation. Our preliminary validation also indicates that gradient direction conflicts between different malicious clients can limit their success. To address this issue, we propose DMBA, a robust method that maintains effectiveness in distributed multi-target backdoor attack scenarios.

\subsubsection{Backdoor attacks in FL}

Backdoor attacks in FL typically inject backdoors into local models with the aim of disrupting the global model through aggregation. Bagdasaryan et al.\cite{bagdasaryan2020backdoor} first introduced backdoor attacks in FL using a model replacement method, amplifying malicious updates from compromised clients to enhance the backdoor's impact. Bhagoji et al.\cite{bhagoji2019analyzing} further improved this by manipulating local learning rates prior to convergence and applied an alternating minimization strategy to balance effectiveness and stealth. Later, Xie et al.\cite{xie2019dba} expanded this to a distributed setup, using trigger fragments to enhance stealth and attack success rates, making it more resistant to defenses. Recently, Liu et al.\cite{liu2024beyond} introduced attack persistence as a new metric for backdoor evaluation, proposing a combinatorics-based distributed trigger strategy that aggregates trigger information to increase attack persistence. These studies reveal the vulnerability of FL to adversarial attacks.

While these studies focus on single backdoor attacks with shared triggers and labels, real-world FL scenarios typically involve multiple attackers with different triggers, labels, and attack timings. To address this, we propose a more challenging threat model and a more practical attack, DMBA. Wang et al.\cite{wang2024dual} introduced a multi-target backdoor attack in FL, but it remains centralized attack settings, assuming a shared trigger and target label across malicious clients. In contrast, DMBA overcomes these limitations and performs well in distributed attack settings.

\subsection{Backdoor defenses in FL} 
\label{2.2}
To mitigate the risks associated with backdoor attacks in FL paradigms, a range of defense mechanisms have been proposed, which can be categorized based on their operational phases into Pre-Aggregation Defenses (Pre-AD), In-Aggregation Defenses (In-AD), and Post-Aggregation Defenses (Post-AD).

Pre-AD methods aim to detect and exclude malicious clients by analyzing update anomalies before aggregation. For example, FoolsGold\cite{fung1808mitigating} detects adversarial clients by examining the diversity of their updates. It requires neither attacker proportion assumptions nor additional information, and works under various client data distributions for backdoor detection. Similarly, FLDetector\cite{zhang2022fldetector} identifies malicious clients by leveraging inconsistencies in their updates. It combines the Cauchy mean theorem with K-means clustering to detect backdoors based on inconsistent updates made by the attacker in iterations. However, Xu et al.\cite{xu2022byzantine} show that some well-crafted backdoor attacks can bypass such defenses. To address this issue, they propose SignGuard, a method that filters out malicious gradients based on the sign of the gradient vectors, thereby enhancing the robustness of the global model. However, these methods often conflict with FL's privacy goals and struggle with frequent data changes, causing false positives or negatives.

In-AD methods mitigate the impact of backdoors during model aggregation by incorporating robust aggregation mechanisms. For example, Differential Privacy (DP)-based techniques reduce backdoor influence by adding noise: Geyer et al.\cite{geyer2017differentially} applied client-level DP through random subsampling and noise injection, while DP-FedAvg\cite{mcmahan2017learning} incorporated noise directly into the aggregation process. Recognizing the limitations of Byzantine-robust methods on non-IID data, Li et al.\cite{li2023experimental} proposed ClippedClustering, which improves clustering-based defenses by adaptively clipping updates. Although these methods help reduce backdoor effects, studies\cite{liu2024beyond}, \cite{wang2024dual} highlight that complete removal remains difficult and often comes at the cost of main task accuracy.

Post-AD methods focus on removing backdoor traces after model aggregation, offering a novel direction in this field. Wu et al.\cite{wu2020mitigating} introduced a federated pruning approach that reduces attack success rates by removing redundant neurons and adjusting extreme weights post-training with minimal accuracy loss. They later proposed a federated unlearning method\cite{wu2022federated}, which removes malicious contributions by subtracting historical updates and restores model performance via knowledge distillation. This method is model-agnostic and does not rely on client participation, enhancing both efficiency and practicality in FL systems.

In Section \ref{5.6}, we assess the robustness of the proposed DMBA method against three advanced defense mechanisms, demonstrating its ability to evade current methods and highlighting the need for more robust defenses in FL.

\section{Problem setting} 
\label{NTM}

\subsection{Analysis of the attack scenario}
\label{3.1}
We adopt the FL-based image classification framework~\cite{mamba2023image} used in DBA~\cite{xie2019dba} as the basic setting. In this framework, a backdoor is typically injected by perturbing selected local training samples with a trigger and modifying their labels so that the local model learns a malicious trigger-target mapping during training~\cite{tolpegin2020data}. The corresponding poisoned updates are then uploaded to the server and aggregated into the global model. To further strengthen the impact of the attack, malicious clients may scale their updates using model replacement techniques~\cite{bagdasaryan2020backdoor}, thereby increasing the contribution of poisoned updates to the aggregated model.

However, practical attack scenarios in FL can be more complex than the conventional single-backdoor setting. In real IoT-based FL systems, a single adversarial entity may compromise multiple distributed devices and control them as malicious clients. More specifically, our setting assumes that:

1) The malicious clients are physically distributed but logically controlled by the same adversary, which is realistic in IoT systems under botnet-based compromise~\cite{kolias2017ddos}, large-scale malware infection~\cite{nguyen2019diot}, or supply-chain attacks~\cite{ladisa2022taxonomy};

2) These clients share a unified global attack objective, which may consist of multiple sub-objectives, i.e., implanting multiple distinguishable backdoors~\cite{gong2022coordinated}, \cite{alharbi2024collusive}, rather than acting as self-interested and competing entities.

Under this setting, the cooperation among malicious clients is inherent and mandatory rather than incentive-driven, and therefore does not rely on trust establishment or negotiation among attackers.

This setting differs fundamentally from a conventional centralized multi-target attack. Although all malicious objectives are coordinated by one adversary, they still have to be executed through separate local datasets, separate local training processes, and separate uploaded updates under the FL protocol. Therefore, the malicious objectives cannot be jointly optimized in a single centralized pipeline, but instead interact mainly through server-side aggregation. Compared with centralized multi-target attacks, this distributed execution pattern is more stealthy at the client level, yet also more difficult to optimize globally because each malicious client contributes only a partial and isolated update to the final model.

\subsection{Motivation}
\label{3.2}

The above setting exposes a practical yet underexplored vulnerability of FL. Existing distributed backdoor attacks, such as DBA, mainly focus on distributing fragments of the same backdoor objective across multiple malicious clients. In contrast, the setting considered in this paper requires the global model to preserve multiple distinct backdoor objectives simultaneously. This difference is crucial because each malicious client is no longer reinforcing the same trigger-target mapping; instead, different malicious clients are optimizing different malicious objectives that may interfere with each other during aggregation.

This interference creates the central challenge of the problem. Since different malicious clients carry different trigger-target mappings, the directions of their uploaded malicious gradients may differ substantially. When these gradients are aggregated by the server, gradient-direction conflicts may arise, which can suppress previously implanted backdoors or even cause them to fail completely. Therefore, simply extending an existing distributed backdoor attack to multiple distinct backdoor objectives is insufficient. The attack must not only implant each backdoor successfully, but also preserve all implanted backdoors jointly in the aggregated global model.

Accordingly, the motivation of this work is to answer the following question: how can a coordinated adversarial entity implant multiple distinct backdoor objectives through separate malicious clients while maintaining the effectiveness of all backdoors after aggregation? To answer this question, the attack design must explicitly address the conflicts among malicious updates during global aggregation and, at the same time, control the interference introduced by learning different backdoor objectives during local malicious training. This observation directly motivates the design of DMBA.

\subsection{Threat model}
\label{3.3}
\textbf{Attack goal:} We consider a coordinated distributed attack setting in which a single adversarial entity controls multiple malicious clients in FL. Each malicious client is assigned one backdoor objective, consisting of a trigger pattern and a predefined target label. The goal of the adversary is to implant all of these backdoor objectives into the global model while preserving the performance of the main task on clean samples. Concretely, the final global model should correctly classify clean inputs, while samples stamped with different triggers should be misclassified into their corresponding target classes.

For the $n$-th malicious client, the local attack objective is defined as:
\begin{align}
\theta_{n}^{*} = \arg\min_{\theta_{n}}  \Bigg( &\sum_{x_{i} \in D_{n}^{cln}} l\left[f\left(x_{i}; \theta_{n}\right), y_{i}\right] \notag \\
 \quad + &\sum_{x_{i^{\prime}} \in D_{n}^{poi}} l\left[f\left(b\left(x_{{i}^{\prime}}; \phi_{n}\right); \theta_{n}\right), \tau_{n}\right] \Bigg) \notag \\
 \label{eq1}
\end{align}
where the first term preserves clean-task performance and the second term enforces the corresponding backdoor objective.

Here, $n =1,2,3,...,N_{mc}$, where $N_{mc}$ is the total number of malicious clients. $\theta_n$ represents the local model parameters of the $n$-th malicious client, and $\theta_{n}^{*}$ denotes the optimized parameters obtained by minimizing the objective function. $D_{n}^{cln}$ denotes the clean local dataset owned by the $n$-th malicious client, $x_{i} \in D_{n}^{cln}$ represents the $i$-th clean training sample in this dataset. $y_i$ is the corresponding ground-truth label of sample $x_i$. $f\left(x_{i}; \theta_{n}\right)$ represents the prediction output of the local model parameterized by $\theta_{n}$ for input $x_i$. $l()$ denotes the cross-entropy loss function. $D_{n}^{poi}$ denotes the poisoned dataset constructed by the $n$-th malicious client, $x_{i^{\prime}} \in D_{n}^{poi}$ represents the $i^{\prime}$-th poisoned sample. $b\left(x_{{i}^{\prime}}; \phi_{n}\right)$ represents the backdoor injection operation applied to sample $x_{{i}^{\prime}}$, where $\phi_{n}$ denotes the trigger pattern associated with the $n$-th malicious client. $\tau_{n}$ denotes the predefined target label corresponding to the $n$-th backdoor. The first term of Eq. \eqref{eq1} ensures that the local model maintains good performance on clean samples, while the second term forces the triggered samples to be classified into the attacker-specified target label.

After local malicious training and server aggregation, the adversary expects all backdoor objectives to be incorporated into the global model. The global objective can be written as:

\begin{align}
\theta_{GM}^{*}=\arg\min_{\theta_{GM}}(
&\sum_{x^{cln}} l[f(x^{cln};\theta_{GM}),y^{cln}] \notag \\+&\sum_n^{N_{mc}}\sum_{x^{poi}}l[f(b(x^{poi};\phi_n);\theta_{GM}),\tau_n]) \notag \\
\label{eq2}
\end{align}
where the global model jointly preserves the clean-task objective and multiple backdoor objectives.

Here, the symbols that have already been defined in Eq. \eqref{eq1} retain the same meanings and are not repeated here for brevity. $\theta_{GM}$ denotes the parameters of the aggregated global model, and $\theta_{GM}^{*}$ represents the optimized global parameters. $x^{cln}$ and $y^{cln}$ denote the clean training samples and their corresponding ground-truth labels used in global aggregation. The summation over $N_{mc}$ ndicates that the global objective aggregates the backdoor objectives from all malicious clients. Compared with Eq.  \eqref{eq1}, Eq. \eqref{eq2} describes the global optimization process, which jointly preserves the main task performance while incorporating multiple backdoor objectives into the aggregated model.

\textbf{Adversary knowledge and capability:} Following Kerckhoffs’ principle~\cite{shannon1949communication} and prior FL backdoor studies~\cite{liu2024beyond}, we assume that the adversarial entity has full control over the compromised clients involved in the attack. Specifically, it can manipulate local data, construct poisoned samples, adjust local training hyperparameters, and determine the attack timing of each malicious client. However, it does not directly control the central server or benign participants. This assumption is practical for IoT-based FL systems, where multiple devices may be compromised through malware propagation, botnet control, or supply-chain attacks, yet must still participate in the FL protocol as separate clients.

\subsection{Preliminary verification}
\label{3.4}
In this section, we conduct preliminary experiments to verify that existing distributed backdoor methods are ineffective under the threat model defined above. The experiments are performed on GTSRB using the basic DBA setting, with adaptations for a distributed multi-target scenario involving three malicious clients. Specifically, the three malicious clients are randomly selected at different attack rounds. Each client is assigned a distinct white-block trigger placed at a different image location (top-left, bottom-left, top-right) and a different target label, while the remaining training protocol follows the standard distributed backdoor setting.

The attack performance, shown in \figref{fig1}, reveals that the global model cannot simultaneously preserve all implanted backdoors. In particular, when a newly assigned backdoor becomes effective, a previously injected one often drops sharply in ASR. This observation indicates that simply extending existing distributed backdoor methods to multiple backdoor objectives is insufficient. The failure is not caused by the lack of poisoning itself, but by the interaction among malicious updates associated with different backdoor objectives.

\textbf{To further analyze this phenomenon}, we record the malicious gradients uploaded by the three malicious clients, denoted as Grad$_1$, Grad$_2$, and Grad$_3$, and compute the cosine similarity between them. Since the task is image classification trained with cross-entropy loss, and to avoid misleading similarity measurements over extremely high-dimensional parameter vectors, we focus on the gradients of the Softmax layer. As reported in \tabref{tab1}, the cosine similarities between different malicious gradients are all negative. This result confirms the existence of severe gradient-direction conflicts during server-side aggregation.

To mitigate this problem, we later introduce the Backdoor Replay (BR) mechanism, which reduces discrepancies among malicious updates before aggregation. However, replaying additional backdoor samples during local training can itself create another source of conflict inside the local model. Therefore, besides aggregation-stage conflicts in the global model, we also examine whether learning different backdoor samples within the same local model introduces gradient inconsistency.

\textbf{Following this idea}, we select three representative trigger strategy variants--Different Transparency Blocks (DTB), Different Channel Masks (DCM), and Different Located Triangles (DLT)--and compare the gradient differences produced when a converged clean model learns different backdoor samples. For each trigger strategy, we compute the gradients corresponding to three backdoors, denoted as grad$_1$, grad$_2$, and grad$_3$, and then calculate their cosine similarities. As shown in \tabref{tab2}, some gradient pairs under DTB and DCM exhibit negative cosine similarity, whereas DLT produces relatively more consistent gradients. These results suggest that trigger design also has a direct influence on local gradient conflicts.

In summary, conventional distributed backdoor training and poorly differentiated trigger strategies can amplify gradient-direction conflicts at both the aggregation stage and the local training stage. This motivates the design of DMBA. Specifically, the BR mechanism is introduced to mitigate conflicts among malicious updates during aggregation, while the proposed Channel-Frequency Composite Trigger (CFCT) strategy is designed to reduce conflicts among different backdoor objectives during local training.

\begin{figure*}[!t]
    \centering 
    \includegraphics[width=0.95\linewidth]{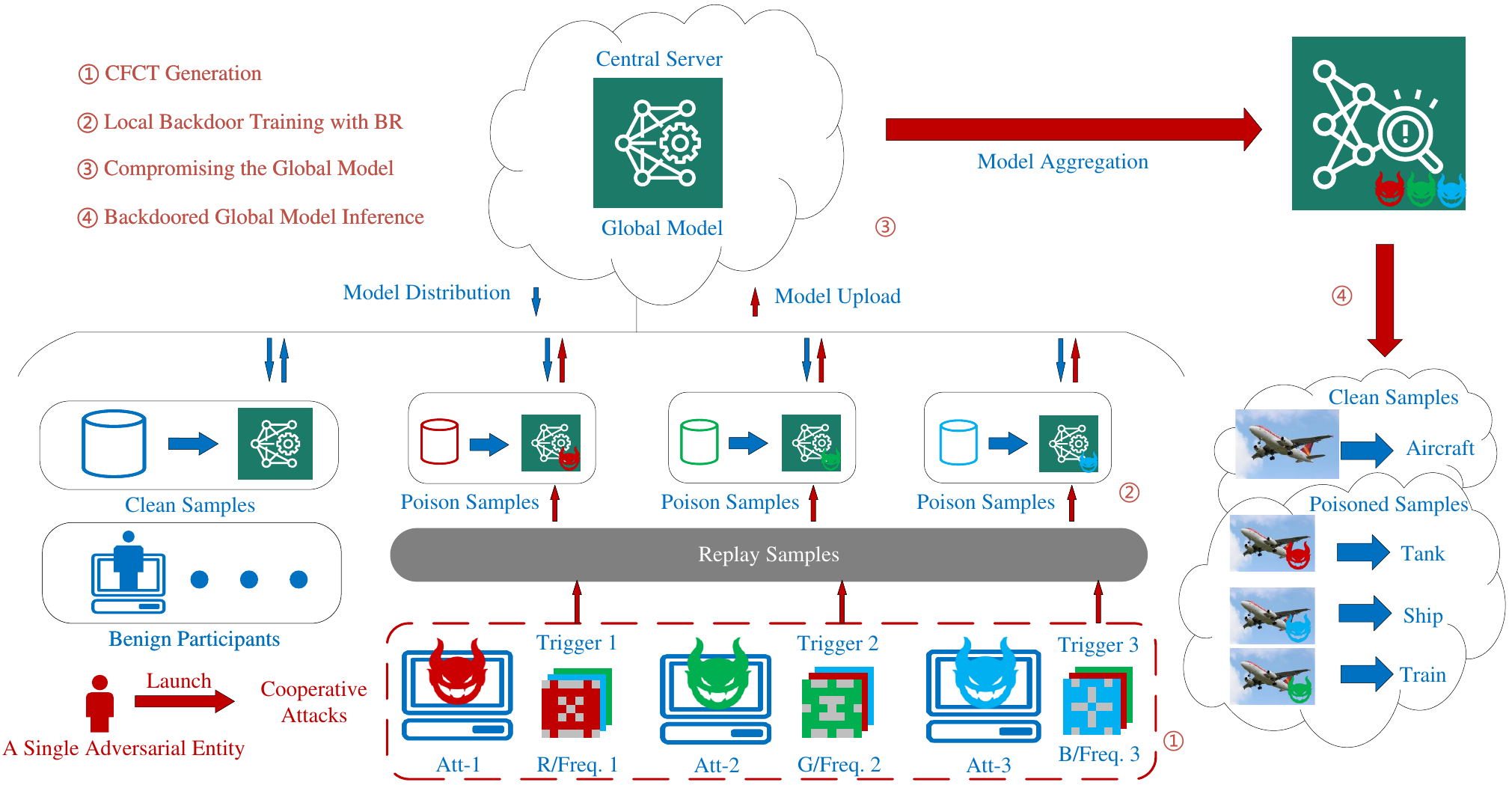} 
    \caption{Workflow of DMBA (example with three malicious clients). (1) CFCT Generation: Composite triggers combining channel and frequency features are crafted and embedded into poisoned samples with flipped labels. (2) Local Backdoor Training with BR: Adversaries train local models using poison samples and replay samples with fine-tuned hyperparameters to maximize backdoor effectiveness. (3) Compromising the Global Model: Malicious model updates are scaled and injected into the aggregation process to implant backdoors into the global model. (4) Backdoored global model inference: The global model maintains clean accuracy while misclassifying triggered samples according to the embedded backdoors.} 
    \label{fig2}
\end{figure*}

\begin{table}[!t]
\begin{center}
\caption{The cosine similarity between malicious gradients submitted by different clients, each embedding a distinct backdoor.}
\label{tab1}
\begin{tabular*}{\columnwidth}{@{\extracolsep{\fill}}@{}c c c c@{}}
\toprule
 & $\text{Grad}_{1}$ vs $\text{Grad}_{2}$
 & $\text{Grad}_{1}$ vs $\text{Grad}_{3}$
 & $\text{Grad}_{2}$ vs $\text{Grad}_{3}$ \\
\midrule
without Replay & \textbf{-0.1124} & \textbf{-0.1677} & \textbf{-0.0438} \\
\bottomrule
\end{tabular*}
\end{center}
\end{table}

This section presents the proposed DMBA framework. We first describe the overall attack workflow in  Section \ref{4.1}, which outlines how coordinated malicious clients inject multiple backdoor objectives into the federated model. Section \ref{4.2} introduces the CFCT strategy designed to improve trigger distinguishability and mitigate local gradient conflicts. Section \ref{4.3} presents the BR mechanism that reduces discrepancies among malicious updates during local training. Finally, Section \ref{4.4} analyzes the computational and memory overhead of the proposed components to demonstrate the scalability of DMBA in practical federated learning settings.

\begin{table}[!t]
\centering
\small
\caption{Gradient differences of the clean model after learning from different backdoor samples. (DTB: Different Transparency Blocks; DCM: Different Channel Masks; DLT: Different Location Triangles)}
\label{tab2}
\begin{tabular*}{\columnwidth}{@{\extracolsep{\fill}}c c c c} 
\toprule
 & $\text{grad}_{1} \text{ vs } \text{grad}_{2}$ 
 & $\text{grad}_{1} \text{ vs } \text{grad}_{3}$ 
 & $\text{grad}_{2} \text{ vs } \text{grad}_{3}$ \\
\midrule
DTB & \textbf{-0.0562} & 0.0832 & 0.0147 \\
DCM & 0.0369 & \textbf{-0.0036} & 0.0703 \\
DLT & 0.0475 & 0.0934 & 0.0477 \\
\midrule 
\end{tabular*}
\end{table}

\section{Distributed multi-target backdoor attack} 
\label{DMBA}

\subsection{Workflow of DMBA}
\label{4.1}
The proposed DMBA is designed for the coordinated distributed threat model defined in Section \ref{NTM}, where a single adversarial entity controls multiple malicious clients and assigns each of them a distinct backdoor objective. \figref{fig2} illustrates the overall workflow of DMBA. The attack consists of four steps, which are described below.


\textbf{Step 1: CFCT generation.} 
After the malicious clients are determined, the adversarial entity assigns each malicious client a distinct trigger-target objective and constructs poisoned samples accordingly. Specifically, composite triggers are generated based on both channel and frequency information to activate different backdoors in a covert and effective manner  (see  Section \ref{4.2} for details). The generated triggers are then embedded into a subset of local training samples, whose labels are flipped to the corresponding target classes. This process produces multiple types of poisoned samples for subsequent local malicious training.

\textbf{Step 2: Local backdoor training with BR.} 
Using the poisoned samples generated in Step 1, each malicious client performs local backdoor training under the guidance of the BR mechanism. Unlike cooperation among self-interested attackers, BR is carried out under unified adversarial control: replay samples associated with other malicious objectives are pre-arranged and incorporated into local malicious training in a coordinated manner. Following prior work~\cite{xie2019dba}, we adopt a smaller learning rate, learning-rate decay, and more local iterations for poisoned samples so as to enhance the backdoor effect in local models.

\textbf{Step 3: Compromising the global model.} 
After local training, the server collects model updates from participating clients and aggregates a selected subset of them to update the global model. If malicious clients are included in the aggregation round, their poisoned updates may be incorporated into the global model, thereby injecting the corresponding backdoor objectives. To further strengthen the malicious effect, we adopt the model replacement method in~\cite{bagdasaryan2020backdoor} and adjust the scaling factor of malicious updates before aggregation. 

\textbf{Step 4: Backdoored global model inference.} 
The final global model is expected to preserve the performance of the main task while simultaneously embedding multiple distinguishable backdoors. In particular, the model should correctly classify clean samples while producing the corresponding adversary-specified target predictions for triggered samples associated with different malicious objectives.

\subsection{Channel-frequency composite trigger}
\label{4.2}
As discussed in Section \ref{3.2}, when different malicious clients are assigned distinct backdoor objectives, learning backdoor samples with similar triggers but different target labels may induce severe local gradient conflicts. To mitigate this problem, we propose a carefully designed multi-backdoor trigger strategy, named Channel-Frequency Composite Trigger (CFCT), to improve the distinguishability among different triggers. CFCT serves as a compensatory design for the distributed multi-target setting by reducing local interference among different backdoor objectives.

According to prior studies~\cite{xue2022imperceptible},\cite{wang2022invisible}, backdoor triggers can generally be designed in either the spatial domain or the frequency domain. Based on this observation, CFCT jointly exploits both domains to reduce feature overlap among different triggers. From the spatial-domain perspective, we adopt a simple yet effective strategy by assigning different color channels (R/G/B) to different malicious clients. From the frequency-domain perspective, we perturb different frequency blocks to construct distinct trigger patterns~\cite{yu2023backdoor}. By separating trigger information across both channel and frequency dimensions, CFCT reduces inter-trigger similarity and thereby alleviates local gradient conflicts.

As previously mentioned training a model on backdoor samples with similar triggers but different target labels often results in severe gradient conflicts. To address this issue, we propose a carefully designed multi-backdoor trigger strategy aimed at improving the distinguishability between triggers. According to prior studies\cite{xue2022imperceptible},~\cite{wang2022invisible}, triggers can generally be categorized into two domains: spatial and frequency. Based on this, we design a CFCT strategy, which leverages both pixel and frequency features to reduce overlap of trigger information across these domains, thereby minimizing inter-trigger similarity. From the spatial domain perspective, we adopt a simple yet widely used method—embedding triggers in different color channels (R/G/B). From the frequency domain perspective, we introduce perturbations in different frequency bands or blocks to serve as triggers\cite{yu2023backdoor}. Further technical details are provided in the following paragraph.

Frequency-domain triggers~\cite{feng2022fiba} have recently attracted attention because they can improve the stealthiness of backdoor attacks by embedding perturbations that are less perceptible to human observers. The basic idea is to transform an image from the spatial domain to the frequency domain and implant the trigger by modifying selected frequency components. This type of trigger is effective because small perturbations in the frequency domain can reliably activate a backdoor while remaining visually subtle. In addition, frequency-domain backdoors often exhibit stronger resistance to conventional defenses. In this work, we adopt the Discrete Cosine Transform (DCT) technique~\cite{wang2022invisible} to transform the pixel matrix of a selected color channel into the frequency domain. To maintain visual imperceptibility, perturbations are injected into relatively small frequency blocks. Compared with other transformation methods, DCT provides favorable energy compaction and smoother signal boundaries, and is less sensitive to high-frequency components, thereby reducing visible artifacts~\cite{zeng2021rethinking}, \cite{hou2023stealthy}.

\begin{figure*}[!t]
    \centering 
    \includegraphics[width=1\linewidth]{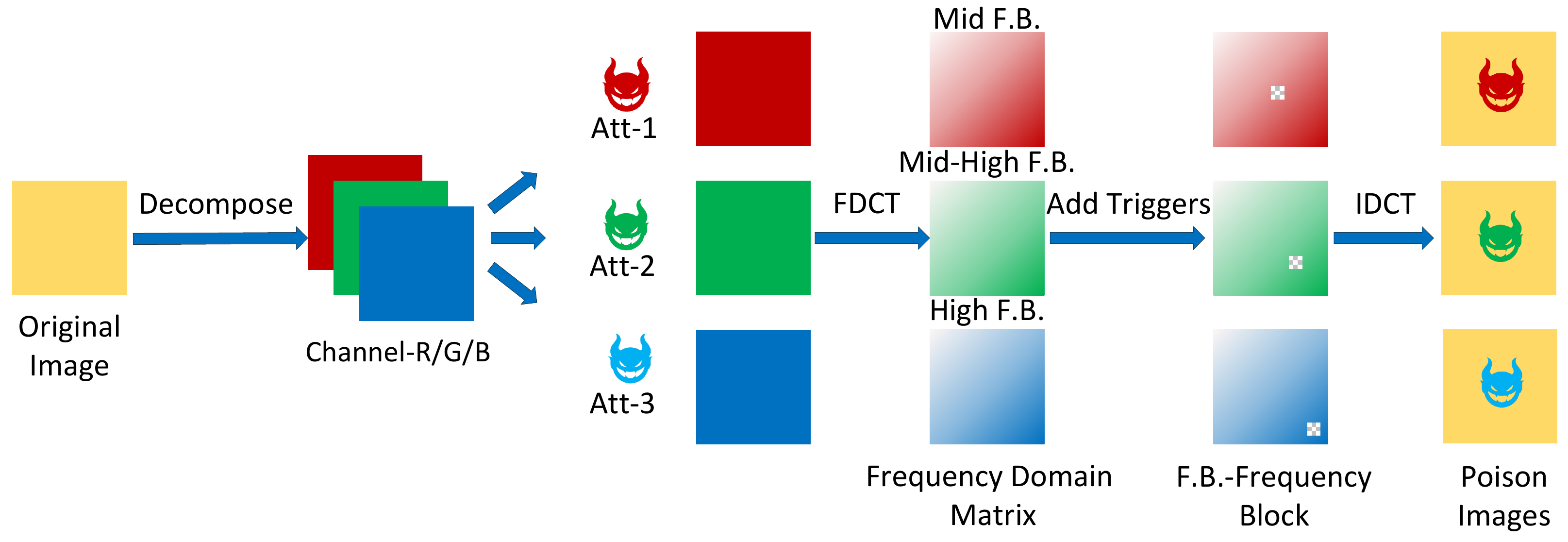} 
    \caption{Illustration of the CFCT strategy (example with three malicious clients). Each attacker independently transforms the pixel matrix of a specific channel into the frequency domain and perturbs distinct frequency blocks. These modifications serve as triggers to activate backdoors associated with different target labels.} 
    \label{fig3}
\end{figure*}

An intuitive illustration of CFCT is shown in \figref{fig3}. For clarity, we use a three-client example. Each malicious client, under the coordinated control of the same adversarial entity, selects a different color channel (R/G/B) and applies DCT to transform that channel into the frequency domain. Different frequency blocks are then perturbed to serve as distinct triggers. Afterward, the modified frequency matrices are transformed back to the spatial domain, and the labels of the corresponding samples are flipped to their assigned target classes, thereby completing the construction of poisoned samples. In our design, perturbations are applied mainly to the mid-to-high frequency region. This choice reduces visible artifacts introduced by high-frequency perturbations while also avoiding the conspicuous distortions that often arise from modifying low-frequency components~\cite{xiao2022invisible}.

\subsection{Backdoor replay mechanism}
\label{4.3}

As discussed earlier, malicious updates associated with different backdoor objectives may exhibit significant divergence, leading to gradient-direction conflicts during aggregation. In the FL literature, this phenomenon is often related to gradient interference~\cite{bagdasaryan2020backdoor}. In distributed multi-backdoor settings, it is typically reflected by a substantial drop, or even complete failure, in the ASR of previously implanted backdoors once new backdoor objectives are injected. In addition to this aggregation-stage conflict, prior studies~\cite{bagdasaryan2020backdoor},\cite{xie2019dba} have suggested that catastrophic forgetting of backdoors~\cite{lesort2020continual} may further degrade attack effectiveness after malicious injection stops. In particular, as benign updates continue to accumulate, the global model tends to refocus on the main task, gradually weakening earlier backdoor effects~\cite{liu2019end}. In our setting, the primary role of BR is to mitigate malicious gradient conflicts during aggregation, while the alleviation of catastrophic forgetting is a secondary benefit.

Inspired by experience replay in Deep Q-Networks~\cite{hessel2018rainbow}, we design a local malicious training method called Backdoor Replay (BR) to improve the joint effectiveness of multiple backdoor objectives in the aggregated global model. Under unified adversarial coordination, each malicious client replays a small number of backdoor samples associated with other malicious objectives during local training. In this way, the local model of each malicious client does not learn only its own backdoor objective. Instead, it is exposed in advance to parameter shifts induced by other backdoor objectives, which helps reduce discrepancies among malicious updates uploaded by different malicious clients. Meanwhile, because replay introduces additional backdoor objectives into local training, it may also create local interference, which is why CFCT is further required.

The BR mechanism consists of the following two steps.

\textbf{(1) Constructing the backdoor replay pool.} Before or during the compromise process, the adversarial entity prepares replay samples for all malicious objectives. Let $D_{br\_n}$ denote the replay samples associated with the $n$-th malicious objective. These replay samples are combined into a shared replay pool, denoted by:
 \begin{equation}
Pool_{br} = \Sigma D_{br\_n}
\label{eq3}
\end{equation}

This pool is maintained under unified control rather than through voluntary exchange among independent attackers.

\textbf{(2) Sharing the replay pool and replaying backdoor samples.} Once the malicious clients are determined, the replay pool $Pool_{br}$ is logically shared among them and incorporated into local malicious training in a coordinated manner. During normal training rounds, replay samples are not used. During attack-triggered rounds, however, each malicious client replays a small number of backdoor samples associated with other malicious objectives. This process is centrally orchestrated by the adversarial entity and does not assume any trust establishment or negotiation among attackers.

For ease of derivation, we first consider the case with three malicious clients. Without loss of generality, let $D_{b1}$ denote the local training set of malicious client 1. This set consists of the original local data $D_{local}$ and a subset of replay samples selected from $Pool_{br}$. The replay samples corresponding to malicious client 1's own backdoor objective, denoted as $D_{br\_1}$, are excluded, and only replay samples associated with the other two malicious objectives are retained. Since $D_{local}$ can be divided into clean samples $D_{cln}$ and local poisoned samples $D_{poi\_1}$, the local training set is derived as follows:

\begin{align}
    D_{b1} &= D_{local} + Pool_{br} \setminus D_{br\_1}  \notag \\
        &= D_{local} + D_{br\_2} + D_{br\_3}  \notag \\
        &= D_{poi\_1} +  D_{cln} + D_{br\_2} + D_{br\_3}  \notag \\
        &= D_{poi\_1} + \sum_{m=2}^3 D_{br\_m} + D_{cln}
\label{eq4}
\end{align}

Other malicious clients follow the same procedure. Accordingly, for a general setting with $N_{mc}$ malicious clients, the local training set of malicious client $n$ can be written as:
\begin{equation}
D_{bn} = D_{poi\_n} + \sum_{\substack{m=1 \\ m \neq n}}^{N_{mc}} D_{br\_m} + D_{cln}
\label{eq5}
\end{equation}
where $D_{poi\_n}$ and $D_{cln}$ denote the local poisoned and clean samples of malicious client $n$, espectively, and $D_{br\_m}$ represents replay samples associated with the other $(N_{mc}-1)$ malicious objectives.

Accordingly, in each poisoned batch, the sample composition is given by:

\begin{equation}
bs_{bn} = bs_{poi\_n} + \sum_{\substack{m=1 \\ m \neq n}}^{N_{mc}} bs_{br\_m} + bs_{cln}
\label{eq6}
\end{equation}

Where the number of each sample type is defined as:

\begin{equation}
\left\{
\begin{aligned}
bs_{poi\_n} &= r_{b} \cdot bs, \,\,  n = 1,2,...,N_{mc} \\
bs_{br\_m} &= r_{br} \cdot bs, \,\,  m = 1,2,...,N_{mc} \,\, and \,\, m \neq n \\
bs_{cln} &= bs - bs_{poi\_n} - \sum_{\substack{m=1 \\ m \neq n}}^{N_{mc}} bs_{br\_m}
\end{aligned}
\right.
\label{eq7}
\end{equation}
Here, $bs_{poi\_n}$ denotes the number of local poisoned samples for malicious client $n$, and $bs_{br\_m}$ denotes the number of replayed samples for each external malicious objective. These quantities are controlled by the local poisoning ratio $r_b$ and the replay ratio $r_{br}$, respectively, while the remaining samples in the batch are clean samples.

\subsection{Complexity Analysis of CFCT and BR Components}
\label{4.4}

To evaluate the computational overhead introduced by the proposed CFCT strategy and BR mechanism, we analyze their time and space complexity in comparison with standard federated training.

Let $|\theta|$ denote the total number of model parameters, $bs$ the batch size, $N_{mc}$ the number of malicious clients, and $d=H*W$ the number of pixels in a single image channel.

\subsubsection{Computational Complexity of CFCT}
For each poisoned sample, CFCT applies a DCT to one selected color channel, perturbs a fixed-size frequency block, and performs an inverse DCT transformation. Using a fast DCT implementation, the 2D transformation requires $O(d\log d)$ operations. Since the frequency perturbation is performed on a constant-size block (e.g., 3*3), it incurs only $O(1)$ cost.

Therefore, the computational complexity for generating one poisoned sample is $O(d\log d)$. Assuming each malicious client generates $r_b \cdot b s$ poisoned samples per batch (where $r_b$ denotes the poisoning ratio), the per-round CFCT overhead for all malicious clients is:
\begin{equation}
O\left(N_{mc} \cdot r_b \cdot bs \cdot d \log d\right)
\label{eq8}
\end{equation}

In contrast, the computational complexity of a single batch training step (forward and backward propagation) is approximately:
\begin{equation}
O(bs \cdot|\theta|)
\label{eq9}
\end{equation}

Since modern deep neural networks typically satisfy $\theta \gg N_{m c} \cdot d \log d$, the CFCT preprocessing overhead is asymptotically dominated by the model training cost. Therefore, CFCT introduces only a lightweight computational burden and does not affect the overall training scalability.

\subsubsection{Computational and Space Complexity of BR}

The BR mechanism modifies only the composition of poisoned batches by replaying a small portion of external backdoor samples. Importantly, it does not increase the batch size.

During poisoning rounds, BR involves: 1) Sampling replay instances from a shared replay pool; 2) Replacing a subset of clean samples in the current batch. These operations incur $O(bs)$ time complexity, which is negligible compared to gradient computation. 

Since the batch size remains unchanged, the training complexity per malicious client remains $O(bs \cdot |\theta|)$, and for all malicious clients $O(N_{mc} \cdot bs \cdot |\theta|)$. Thus, BR does not increase the asymptotic time complexity of federated training. Thus, BR does not increase the asymptotic time complexity of federated training.

Regarding space complexity, suppose each malicious client constructs $K$ replay samples. The shared replay pool requires $O(N_{mc} \cdot K \cdot d)$ storage. Additionally, each malicious client temporarily stores replay samples within a mini-batch, requiring at most $O(N_{mc} \cdot r_{br} \cdot bs \cdot d)$, where $r_{br}$ denotes the replay ratio. Including model parameters, the overall space complexity becomes: 
\begin{equation}
O\left(N_{mc} K d+|\theta|\right)
\label{eq10}
\end{equation}

In practical settings, since $K$ and $r_{br}$ are small constants and $|\theta|$ is typically large, the dominant term remains the model parameter storage. Therefore, BR introduces only linear overhead with respect to the number of malicious clients and does not compromise memory scalability.

In summary, CFCT introduces a lightweight $O(d \log d)$ preprocessing cost per poisoned sample, while BR preserves the original training complexity $O(bs \cdot |\theta|)$. Neither component increases the asymptotic computational order of federated learning, ensuring that DMBA remains computationally scalable even under multiple distributed malicious clients.

\section{Experiments \& analyses} 
\label{E&A}

\subsection{Experimental settings}
\label{5.1}
\subsubsection{General setup}
\label{5.1.1}
Experiments were conducted on a server equipped with four NVIDIA A100 GPUs using PyTorch~\cite{paszke1912pytorch} as the software framework. The performance of DMBA was evaluated on three datasets~\cite{krizhevsky2009learning},\cite{stallkamp2011german} and three models~\cite{he2016deep}, as summarized in \tabref{tab3}. Here, \textit{Lw} denotes a lightweight version of ResNet-18, in which the number of channels in all convolutional layers is halved. \textit{Clean ACC} refers to the accuracy of the primary task achieved by the converged global model in
the absence of any attack.

Although CIFAR-10 and CIFAR-100 are general-purpose image classification benchmarks, they have been widely adopted in FL research to simulate distributed perception tasks in IoT environments~\cite{pan2024one}, particularly for edge vision~\cite{bonazzi2025picosam2} and smart sensing applications~\cite{dutto2024collaborative}. These datasets allow controlled evaluation of security threats in large-scale distributed settings. The GTSRB dataset corresponds directly to traffic sign recognition tasks in intelligent transportation systems~\cite{wang2024defending}, which are representative IoT scenarios involving distributed vehicular and roadside devices~\cite{n2025real}. Therefore, our dataset selection reflects typical perception tasks in IoT-based FL systems.

\subsubsection{FL setup}
\label{5.1.2}

The central server adopts the FedAvg aggregation algorithm~\cite{mcmahan2017communication}, with a learning rate of 0.01. Each client performs local training using a batch size of 64, the cross-entropy loss function, and the SGD optimizer\cite{konevcny2016federated}, \cite{bagdasaryan2020backdoor}. Clean samples are trained for 2 epochs with a learning rate of 0.1, while poisoned samples are trained for 6 epochs with an initial learning rate of 0.05, which decays by a factor of 10 every 2 epochs. As discussed in prior FL backdoor studies, this configuration helps strengthen the malicious objective in local models.

Similar to DBA, we simulate non-I.I.D. data distributions in FL by partitioning each dataset with a Dirichlet distribution using $\alpha = 0.8$, esulting in a reasonably heterogeneous client distribution. Without loss of generality, after the main task converges, a single adversarial entity is assumed to coordinate multiple malicious clients and asynchronously inject distinct backdoor objectives into different rounds. To avoid excessive interference among malicious updates, attacks are launched every other round. Specifically, the first malicious client starts the attack at round 202 and performs three attack rounds to strengthen its corresponding backdoor effect. This setting is consistent with the coordinated distributed threat model defined in Section \ref{NTM}.

\begin{table}[!t]
\centering
\small
\caption{Datasets and other settings.}
\label{tab3}
\begin{tabular*}{\columnwidth}{@{\extracolsep{\fill}}c c c c}
\toprule
Datasets      & CIFAR-10         & CIFAR-100        & GTSRB \\
\midrule
Training/Test & 50{,}000/10{,}000 & 50{,}000/10{,}000 & 39{,}209/12{,}569 \\
Models        & (Lw)ResNet-18    & ResNet-18        & 4Conv+2fc \\
Labels        & 10               & 100              & 43 \\
Target Labels & \#0, \#4, \#6    & \#10, \#47, \#59 & \#0, \#20, \#29 \\
Clean ACC     & 77\%             & 72\%             & 99\% \\
\midrule
\end{tabular*}
\end{table}

\subsubsection{CFCT setup}
\label{5.1.3}
As described in Section \ref{4.3} three malicious clients, denoted as Att-1, Att-2, and Att-3 for convenience, are assigned one of the R/G/B channels, respectively, to embed their triggers. Each selected channel is transformed into the frequency domain using DCT, and perturbations are injected into the mid-to-high frequency region. Specifically, the starting positions of the perturbed frequency blocks are set to $f=$ (15,15), (20,20), and (25,25). An important refinement is that we use low-amplitude multi-point perturbations, i.e., frequency blocks, instead of the traditional single-point perturbation. This strategy allows the perturbation energy to be distributed more smoothly in the frequency domain, which improves stealthiness without significantly sacrificing attack effectiveness~\cite{khayam2003discrete}. In addition, the attack no longer depends on a single frequency component, which enhances its resilience against certain defenses. In our implementation, we perturb 3*3 frequency blocks with an amplitude of +100, and the perturbed frequency-domain matrices are then transformed back to the spatial domain by inverse DCT to complete trigger construction.

\subsubsection{BR setup}
\label{5.1.4}

Based on preliminary experience and repeated experiments, we set the poisoning rate of local poisoned samples to $r_b=8/64$, and the replay rate of each external malicious objective to $r_{br}=3/64$. Under this configuration, both local poisoned samples and replayed backdoor samples achieve strong performance. To simplify implementation without affecting the essence of the attack design, we adopt a direct replacement method, referred to as \textit{DirectRep}, instead of the original shared-pool realization, denoted as \textit{ReplayPool}.

Taking Att-1 as an example, during poisoning rounds, it randomly selects $2r_{br}\cdot bs_n$ clean samples from the current batch and replaces them with $r_{br}\cdot bs_n$ replay samples associated with the malicious objectives of Att-2 and Att-3, respectively. Although the implementation differs from explicitly drawing samples from a physically maintained shared pool, the two mechanisms are functionally equivalent under the unified control of the same adversarial entity, since they replay the same external backdoor samples during local malicious training. Therefore, DirectRep preserves the intended effect of BR while simplifying data handling. As shown in \tabref{tab4}, the two implementations achieve comparable attack performance, which supports the use of DirectRep in our experiments.

\begin{table}[!t]
\centering
\scriptsize
\caption{Comparison of attack performance between ReplayPool and DirectRep replay methods.}
\label{tab4}

\setlength{\tabcolsep}{2.2pt}
\begin{tabular*}{\columnwidth}{@{\extracolsep{\fill}}@{}llcccccc@{}}
\toprule
\multicolumn{2}{l}{Datasets}
 & \multicolumn{2}{c}{CIFAR-10}
 & \multicolumn{2}{c}{CIFAR-100}
 & \multicolumn{2}{c}{GTSRB} \\
\cmidrule(lr){3-4}\cmidrule(lr){5-6}\cmidrule(lr){7-8}
& & \textbf{DirectRep} & ReplayPool
  & \textbf{DirectRep} & ReplayPool
  & \textbf{DirectRep} & ReplayPool \\
\midrule
\multicolumn{2}{l}{ACC}
 & 73.30\% & 72.80\%
 & 52.63\% & 52.93\%
 & 98.13\% & 98.07\% \\
\midrule
\multirow{3}{*}{ASR}
 & Att-1 & 96.91\% & 98.92\%
           & 91.19\% & 96.99\%
           & 98.72\% & 93.73\% \\
 & Att-2 & 94.49\% & 99.08\%
           & 95.80\% & 94.99\%
           & 99.83\% & 99.80\% \\
 & Att-3 & 99.28\% & 92.68\%
           & 92.25\% & 95.25\%
           & 99.30\% & 95.88\% \\
\bottomrule
\end{tabular*}
\end{table}

\subsubsection{Baseline method settings}
\label{5.1.5}
Since existing FL backdoor studies mainly focus on trigger design for single backdoor attacks, there is still limited systematic exploration of distributed multi-target backdoor attacks or composite trigger strategies. As a result, there is no widely accepted classical baseline tailored specifically for the threat model studied in this paper. Although prior works have proposed several multi-backdoor trigger patterns~\cite{xue2020one},\cite{wang2022dispersed},\cite{kwon2020targetnet}, they do not explicitly address the distributed multi-target setting considered here, nor do they optimize trigger design for gradient conflict mitigation.

To provide fair and representative comparisons, we construct three baseline methods--DTB, DCM, and DLT--by adapting established trigger design ideas from prior studies. Each baseline preserves the original design philosophy and its key parameter settings as much as possible, while introducing only the minimum modifications necessary to make it applicable to the coordinated distributed multi-target scenario. In this way, the comparison focuses on the relative effectiveness of different trigger-design strategies under the same distributed attack framework.

{\bf DTB} (Triggers with \textit{Different Transparency Blocks}): This method is adapted from the trigger design in~\cite{xue2020one}, where the same fixed pattern with different transparency levels is used to activate different backdoors. To strengthen activation, we adopt the white-block trigger from BadNets~\cite{gu2017badnets} and place it at the four corners of the image. To reduce similarity among different backdoor objectives, we use clearly separated transparency levels of 20\%, 50\%, and 80\% for different malicious objectives.

{\bf DCM} (Triggers with \textit{Different Channel Masks}): Inspired by the dispersed pixel perturbation trigger in~\cite{wang2022dispersed}, this method decomposes the original three-dimensional binary mask into three one-dimensional binary masks and applies them separately to the R/G/B channels to activate different backdoors. This design reduces trigger overlap across malicious objectives while preserving both effectiveness and stealthiness.

{\bf DLT} (Triggers with \textit{Differently Located Triangles}): This method follows the trigger design in TargetNet~\cite{kwon2020targetnet}, where fixed patterns placed at different image locations are used to activate different target backdoors. The goal is to keep the overall perturbation magnitude at a similar level across images, thereby ensuring a relatively consistent attack intensity.

\subsubsection{Evaluation metrics}
\label{5.1.6}

We employ several metrics to comprehensively evaluate DMBA. Accuracy (ACC) measures the impact of the attack on the main task, and ASR directly reflects the strength of the backdoor attack~\cite{li2023poisoning}. However, because DMBA involves multiple distinct backdoor objectives, we further use ASR$_{n}$ to evaluate the attack success rate of the $n$-th malicious objective individually. As noted in~\cite{liu2024beyond}, due to the continual-learning nature of FL, a strong backdoor should remain effective for some time even after malicious injection stops. Therefore, we additionally introduce ASR$_{n}$-30 to measure attack persistence 30 rounds after the corresponding attack stops. To evaluate trigger stealthiness, we use two perceptual similarity metrics: \textit{SSIM} and \textit{LPIPS}. Their detailed definitions are given below.

{\bf ACC}$\uparrow$\cite{jiang2023color}: The classification accuracy of the global model on clean samples after the attack is deployed. A higher ACC indicates less degradation of the main task and thus better attack concealment.

{\bf ASR$_{n}$}$\uparrow$ and \textbf{ASR$_{n}$-30}$\uparrow$: The probability that triggered samples associated with the $n$-th malicious objective are misclassified into the designated target class, measured at the round when the attack stops and 30 rounds later, respectively. Higher values indicate stronger and more persistent attack effectiveness.

{\bf SSIM}$\uparrow$\cite{wang2004image}: SSIM evaluates the perceptual similarity between two images by comparing luminance, contrast, and structural information. A higher SSIM value indicates that the poisoned sample remains structurally closer to the clean sample, implying better stealthiness.

{\bf LPIPS}$\downarrow$\cite{zhang2018unreasonable}: LPIPS measures perceptual similarity using deep feature representations extracted by a pretrained neural network. A lower LPIPS value indicates that the poisoned sample is perceptually closer to the original sample, which also implies better stealthiness.

\begin{table}[!t]
\centering
\scriptsize
\caption{Comparison of DMBA performance with and without BR mechanism on various datasets.}
\label{tab5}

\setlength{\tabcolsep}{3pt}

\begin{tabular*}{\columnwidth}{@{\extracolsep{\fill}} cccccccc@{}}
\toprule
\multicolumn{2}{c}{\textbf{Datasets}}
  & \multicolumn{2}{c}{CIFAR-10}
  & \multicolumn{2}{c}{CIFAR-100}
  & \multicolumn{2}{c}{GTSRB} \\
\cmidrule(lr){3-4}\cmidrule(lr){5-6}\cmidrule(lr){7-8}
& & w/ BR & \textbf{w/o BR}
  & w/ BR & \textbf{w/o BR}
  & w/ BR & \textbf{w/o BR} \\
\midrule
\multicolumn{2}{c}{ACC}
  & 73.30\% & 69.38\%
  & 52.63\% & 45.01\%
  & 98.13\% & 94.70\% \\
\midrule
\multirow{3}{*}{ASR}
 & Att-1 & 96.91\% & \textbf{7.30\%}
         & 91.19\% & \textbf{0\%}
         & 98.72\% & \textbf{0\%} \\
 & Att-2 & 94.49\% & 99.98\%
         & 95.80\% & 99.96\%
         & 99.83\% & 99.96\% \\
 & Att-3 & 99.28\% & \textbf{3.58\%}
         & 92.25\% & \textbf{0\%}
         & 99.30\% & \textbf{0\%} \\
\midrule
\multirow{3}{*}{ASR-30}
 & Att-1 & 78.42\% & \textbf{3.82\%}
         & 69.07\% & \textbf{0\%}
         & 96.79\% & \textbf{0\%} \\
 & Att-2 & 89.34\% & 94.99\%
         & 87.96\% & 97.30\%
         & 99.67\% & 97.22\% \\
 & Att-3 & 97.01\% & \textbf{38.88\%}
         & 58.71\% & \textbf{0\%}
         & 99.25\% & \textbf{0\%} \\
\bottomrule
\end{tabular*}
\end{table}

\subsection{Ablation study of BR mechanism}
\label{5.2}
To highlight the critical role of the BR mechanism in DMBA, we conduct an ablation study that isolates its contribution under the coordinated distributed multi-target setting. The experimental settings are as follows.

{\bf (1) Complete Model} (DMBA with BR mechanism): 
As described in Section \ref{5.1}, each poisoned batch contains 64 samples, including 8 local poisoned samples, 3 replayed backdoor samples associated with each of the two external malicious objectives, and 50 clean samples.

{\bf (2) Ablation Group} (DMBA without BR mechanism): 
In this setting, each poisoned batch consists of 14 local poisoned samples and 50 clean samples, without any replayed backdoor samples. To ensure a fair comparison, the total number of poisoned samples in each batch is kept unchanged. In this way, the overall attack cost remains comparable, and the effect of the BR mechanism can be isolated more clearly.

The experimental results are reported in \tabref{tab5}. It can be observed that BR substantially improves the overall attack performance of DMBA. In the ablation group without BR, only the most recently injected malicious objective remains effective, whereas the other backdoors almost completely fail, with ASR values close to zero on several datasets. In contrast, the complete model with BR consistently maintains strong attack performance for all malicious objectives, and the average ASR remains above 83\% even 30 rounds after the attacks stop. These results indicate that replaying a small number of external backdoor samples is essential for preserving multiple backdoor objectives jointly in the aggregated global model.

\begin{figure}[!t]
    \centering 
    \includegraphics[width=1\linewidth]{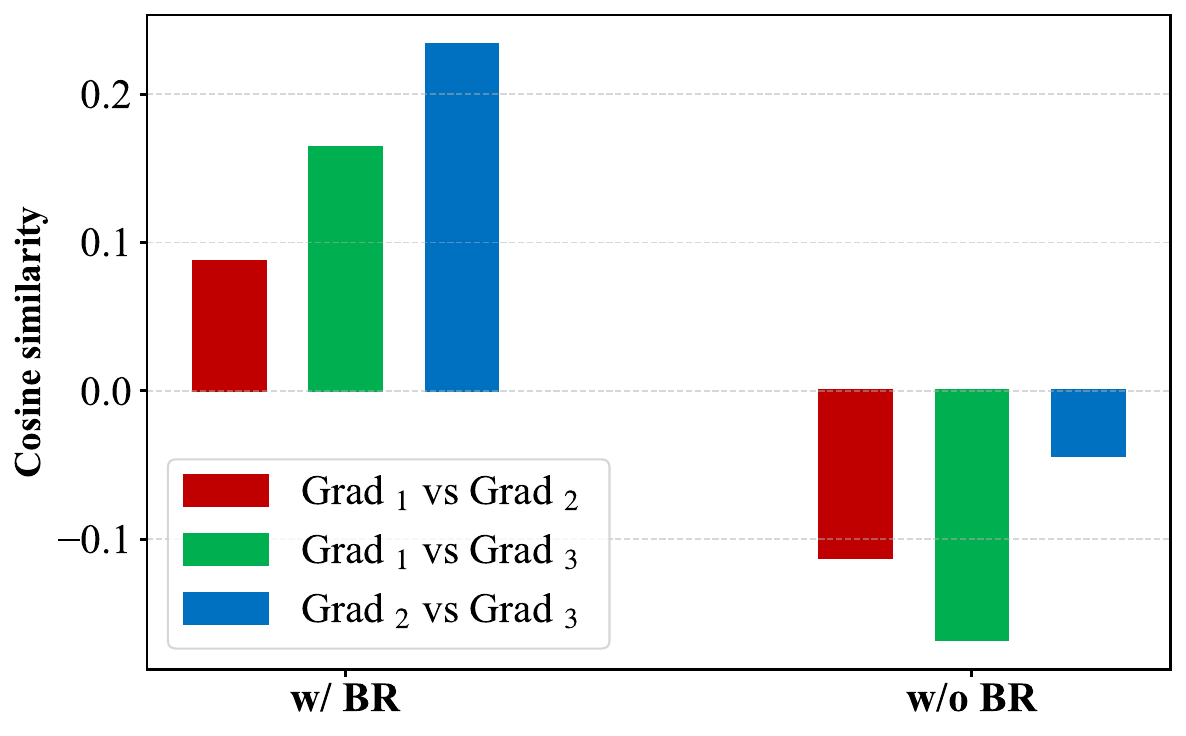} 
    \caption{Comparison of gradient differences from different malicious clients with and without the BR mechanism.} 
    \label{fig4}
\end{figure}

To further explain this improvement, we follow the analysis protocol in Section \ref{3.4} and record the gradients uploaded by the three malicious clients in both settings, denoted as Grad$_1$, Grad$_2$, and Grad$_3$. The cosine similarities between these gradients are shown in \figref{fig4}. In the ablation group, all pairwise similarities are negative, indicating severe gradient-direction conflicts among malicious updates. By contrast, after BR is introduced, the corresponding cosine similarities become consistently positive. This change provides direct evidence that BR effectively reduces discrepancies among malicious updates before aggregation and converts strongly conflicting gradients into more aligned update directions.

Taken together, the results in \tabref{tab5} and \figref{fig4} verify the design motivation of BR from both performance and optimization perspectives. Without BR, distinct malicious objectives interfere with one another so strongly that previously implanted backdoors are quickly suppressed. With BR, each malicious client is exposed in advance to a small amount of replayed information from other malicious objectives, which alleviates aggregation-stage gradient conflicts and enables the global model to preserve multiple backdoors simultaneously. This observation is consistent with the threat model and motivation presented in Section \ref{NTM}.

\subsection{Main results}
\label{5.3}
This section evaluates the overall performance of DMBA with BR across different datasets, focusing on the effectiveness and stealthiness of each implanted backdoor. To ensure reliability, all experiments were repeated three times, and the average results are reported in \tabref{tab6}, where “Avg.” denotes the average performance over the three malicious objectives. As shown in the table, DMBA consistently achieves strong attack performance under the coordinated distributed multi-target setting. In particular, the ASRs of individual backdoors remain high at the end of the attack, and the average ASR for all backdoors stays above 92\% across the tested datasets. Even 30 rounds after the attack stops, the average ASR remains above 83\%, indicating that DMBA is able not only to implant multiple distinct backdoors, but also to preserve them jointly in the aggregated global model for a sustained period.

In terms of trigger stealthiness, DMBA also shows favorable performance. As reported in \tabref{tab6}, the proposed CFCT strategy achieves the highest SSIM values and among the lowest LPIPS values on most settings, indicating that the generated poisoned samples remain highly similar to the corresponding clean samples in both structural and perceptual senses. This observation is further supported by the visual examples in \tabref{tab7}, where the differences between the poisoned and clean samples produced by DMBA are barely perceptible. These results suggest that DMBA offers a strong balance between attack effectiveness and visual concealment.

\begin{figure}[!t]
    \centering 
    \includegraphics[width=1\linewidth]{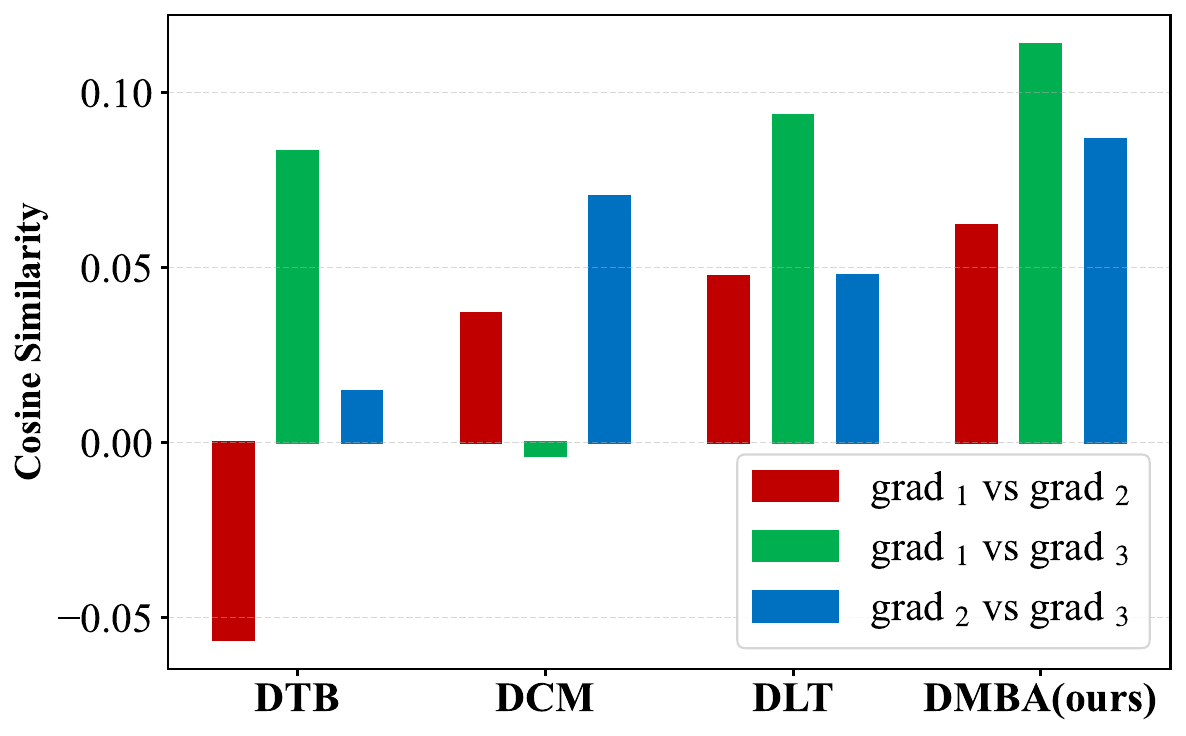} 
    \caption{Comparison of gradient differences in the same model after learning from different backdoor samples under different trigger strategies.} 
    \label{fig5}
\end{figure}

Compared with the baseline methods, DMBA exhibits more stable overall performance across different malicious objectives. Although DLT can also achieve strong ASR values in some settings, its trigger patterns are visibly more noticeable than those of DMBA, as reflected by both the quantitative metrics in \tabref{tab6} and the sample visualizations in \tabref{tab7}. By contrast, some backdoors under DTB and DCM become highly unstable, and in several cases their ASRs drop sharply for individual malicious objectives. These comparisons indicate that achieving high average attack success alone is insufficient in the distributed multi-target setting; rather, the key requirement is to maintain all implanted backdoors simultaneously while preserving acceptable stealthiness. DMBA is the method that best satisfies this requirement among the compared approaches.

To further understand why DMBA performs better, we analyze the gradient-direction conflicts introduced by different trigger strategies during local training. Following the procedure in Section \ref{3.4}, for each trigger strategy we select a converged clean global model and compute the gradients after training on 14 poisoned samples for each of the three malicious objectives, denoted as grad$_1$, grad$_2$, and grad$_3$. We then calculate the cosine similarity between each pair of gradients, and the results are shown in \figref{fig5}. Compared with DTB, DCM, and DLT, the CFCT strategy consistently yields positive and larger cosine similarities across all gradient pairs. This result indicates that CFCT more effectively reduces local gradient interference among distinct malicious objectives, which directly supports the design rationale described in Section \ref{4.2}.

Overall, the results demonstrate that DMBA succeeds not merely because it uses a stealthy trigger or a stronger poisoning strategy, but because its two components address the two core difficulties of the coordinated distributed multi-target setting in a complementary manner. Specifically, BR mitigates aggregation-stage conflicts among malicious updates, while CFCT reduces local interference among different backdoor objectives. Their combination enables the global model to retain multiple distinguishable backdoors with both strong effectiveness and strong concealment.

\begin{table*}[!t]
\centering
\scriptsize
\caption{Comparing attack and stealth performance of different attacks on various datasets.}
\label{tab6}

\resizebox{\textwidth}{!}{
\begin{tabular}{llcccccccccccc}
\toprule
\multicolumn{2}{c}{Datasets}
  & \multicolumn{4}{c}{CIFAR-10}
  & \multicolumn{4}{c}{CIFAR-100}
  & \multicolumn{4}{c}{GTSRB} \\
\cmidrule(lr){3-6}\cmidrule(lr){7-10}\cmidrule(lr){11-14}
\multicolumn{2}{c}{Methods}
  & DTB & DCM & DLT & DMBA
  & DTB & DCM & DLT & DMBA
  & DTB & DCM & DLT & DMBA \\
\midrule
\multicolumn{2}{c}{Acc}
  & 71.86\% & 71.02\% & 73.16\% & 72.97\%
  & 55.83\% & 52.47\% & 56.32\% & 53.74\%
  & 98.11\% & 99.47\% & 98.33\% & 97.72\% \\
\midrule
\multirow{4}{*}{ASR$\uparrow$}
 & Att-1 & 81.57\% & 88.63\% & 96.32\% & 96.73\%
         & 93.26\% & 80.88\% & 99.83\% & 86.68\%
         & 3.37\%  & 65.72\% & 99.36\% & 94.51\% \\
 & Att-2 & 84.17\% & 97.10\% & 95.37\% & 93.97\%
         & 94.57\% & 91.08\% & 99.67\% & 96.19\%
         & 100\%   & 96.63\% & 95.41\% & 91.76\% \\
 & Att-3 & 91.06\% & 95.69\% & 92.83\% & 97.06\%
         & 99.29\% & 23.65\% & 99.04\% & 80.42\%
         & 91.73\% & 92.30\% & 89.31\% & 95.09\% \\
\cdashline{2-14}\addlinespace[0.4ex]
 & Avg.  & 85.60\% & 93.81\% & 94.84\% & \textbf{95.92\%}
         & 95.71\% & 65.20\% & \textbf{99.51\%} & 87.76\%
         & 65.03\% & 84.88\% & \textbf{94.69\%} & 93.79\% \\
\midrule
\multirow{4}{*}{ASR-30$\uparrow$}
 & Att-1 & 28.08\% & 60.98\% & 87.09\% & 73.12\%
         & 59.39\% & 65.72\% & 99.36\% & 71.40\%
         & 89.92\% & 86.47\% & 84.63\% & 93.14\% \\
 & Att-2 & 74.73\% & 90.53\% & 91.51\% & 88.84\%
         & 82.47\% & 75.82\% & 93.24\% & 84.45\%
         & 99.97\% & 88.71\% & 92.78\% & 99.67\% \\
 & Att-3 & 92.43\% & 83.92\% & 87.39\% & 95.99\%
         & 98.40\% & 19.77\% & 98.97\% & 67.76\%
         & 51.18\% & 88.05\% & 86.59\% & 78.07\% \\
\cdashline{2-14}\addlinespace[0.4ex]
 & Avg.  & 65.08\% & 78.48\% & \textbf{88.66\%} & 85.98\%
         & 80.09\% & 53.77\% & \textbf{97.19\%} & 74.54\%
         & 80.36\% & 87.74\% & 88.00\% & \textbf{90.29\%} \\
\midrule
\multirow{4}{*}{SSIM$\uparrow$}
 & Att-1 & 0.95843 & 0.95418 & 0.97250 & \textbf{0.98446}
         & 0.95852 & 0.95216 & 0.97392 & \textbf{0.98428}
         & 0.93283 & 0.91078 & 0.96460 & \textbf{0.97694} \\
 & Att-2 & 0.93318 & 0.95265 & 0.97164 & \textbf{0.98396}
         & 0.93533 & 0.95073 & 0.97217 & \textbf{0.98420}
         & 0.90495 & 0.91425 & 0.96187 & \textbf{0.97697} \\
 & Att-3 & 0.91964 & 0.95073 & 0.97262 & \textbf{0.98328}
         & 0.92287 & 0.94774 & 0.97401 & \textbf{0.98343}
         & 0.89517 & 0.91729 & 0.96463 & \textbf{0.97792} \\
\cdashline{2-14}\addlinespace[0.4ex]
 & Avg.  & 0.93708 & 0.95252 & 0.97225 & \textbf{0.98390}
         & 0.93891 & 0.95021 & 0.97337 & \textbf{0.98397}
         & 0.91098 & 0.91411 & 0.96370 & \textbf{0.97728} \\
\midrule
\multirow{4}{*}{LPIPS$\downarrow$}
 & Att-1 & 0.0784 & 0.0586 & 0.0675 & \textbf{0.0283}
         & 0.0746 & 0.0594 & 0.0630 & \textbf{0.0266}
         & 0.1174 & 0.1247 & 0.0953 & \textbf{0.0505} \\
 & Att-2 & 0.1251 & 0.0950 & 0.0668 & \textbf{0.0445}
         & 0.1176 & 0.0918 & 0.0638 & \textbf{0.0489}
         & 0.1721 & 0.1905 & 0.0867 & \textbf{0.0831} \\
 & Att-3 & 0.1553 & 0.0321 & 0.0684 & \textbf{0.0074}
         & 0.1460 & 0.0320 & 0.0631 & \textbf{0.0116}
         & 0.2018 & 0.0621 & 0.0915 & \textbf{0.0248} \\
\cdashline{2-14}\addlinespace[0.4ex]
 & Avg.  & 0.1196 & 0.0619 & 0.0676 & \textbf{0.0267}
         & 0.1127 & 0.0611 & 0.0633 & \textbf{0.0290}
         & 0.1638 & 0.1258 & 0.0912 & \textbf{0.0528} \\
\bottomrule
\end{tabular}
} 
\end{table*}

\begin{table}[!t]
    \centering
    \footnotesize
    \caption{Examples of backdoor samples with different trigger strategies.}
    \label{tab7}
    \setlength{\tabcolsep}{2pt} 
    \begin{tabular*}{\columnwidth}{@{\extracolsep{\fill}}
        >{\centering\arraybackslash}m{1.0cm}  
        >{\centering\arraybackslash}m{0.6cm}  
        >{\centering\arraybackslash}m{1.5cm}  
        >{\centering\arraybackslash}m{1.5cm}  
        >{\centering\arraybackslash}m{1.5cm}  
        >{\centering\arraybackslash}m{1.5cm}  
    }
        \toprule
        Clean &   & DTB & DCM & DLT & \textbf{DMBA} \\ 
        \midrule
        \addlinespace[0.8ex]
        & Att-1 
        & \includegraphics[width=0.9cm]{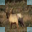} 
        & \includegraphics[width=0.9cm]{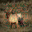} 
        & \includegraphics[width=0.9cm]{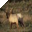} 
        & \includegraphics[width=0.9cm]{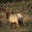} \\
        \addlinespace[0.8ex]
        \includegraphics[width=0.9cm]{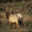} & Att-2 
        & \includegraphics[width=0.9cm]{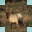} 
        & \includegraphics[width=0.9cm]{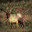} 
        & \includegraphics[width=0.9cm]{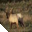} 
        & \includegraphics[width=0.9cm]{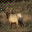} \\
        \addlinespace[0.8ex]
        & Att-3 
        & \includegraphics[width=0.9cm]{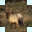} 
        & \includegraphics[width=0.9cm]{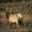} 
        & \includegraphics[width=0.9cm]{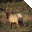} 
        & \includegraphics[width=0.9cm]{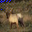} \\
        \bottomrule
    \end{tabular*}
\end{table}

\subsection{Hyperparameter sensitivity analyses}
\label{5.4}
This section analyzes several key factors that influence the performance of DMBA. Unless otherwise specified, only one representative dataset is reported for each group of experiments for brevity, while additional implementation details and auxiliary results are provided in the \textbf{Appendix}.

\subsubsection{Impact of frequency block starting position}
\label{5.4.1}
Motivated by prior studies on frequency-domain backdoors, we first investigate whether the perturbation location affects the attack performance of DMBA. As described in Section \ref{5.1}
each malicious client perturbs a 3*3 block in the frequency domain. Following~\cite{wang2022invisible}, we place this block along the diagonal of the frequency matrix and denote the top-left coordinate of the block by $f$. Accordingly, the starting position of a frequency block is abbreviated as $f$. We experiment with different combinations of $f$ for the three malicious objectives, and the results on CIFAR-10 are reported in \tabref{tab8}.

\begin{table}[!t]
    \centering
    \caption{Effect of frequency block position on attack performance in CIFAR-10.}
    \label{tab8}
    \setlength{\tabcolsep}{2.5mm}{
        \begin{tabular*}{\columnwidth}{@{\extracolsep{\fill}} cc @{\hspace{12.5mm}} ccc}
         \toprule 
          \multicolumn{2}{c}{\hspace*{-7mm}Positions}& 15/20/25 &  15/15/15 &  5/15/25 \\
         \midrule
         \multicolumn{2}{c}{\hspace*{-7mm}ACC} &  73.30\% &  72.66\% &  73.66\%\\
         \midrule
         \multirow{4}{*}[-1ex]{\centering ASR}&  Att-1 &  96.91\% &  76.18\% &  78.86\% \\[1.5pt]
          &  Att-2 &  94.49\% & 91.10\% & 97.79\%\\[1.5pt]
          ~ &  Att-3 &  99.28\% &  97.83\% & 98.07\% \\[1.5pt]
            \cdashline{2-5} 
            \addlinespace[0.4ex]
         ~ & Avg. &  \textbf{96.89\%} &  88.37\% & 91.57\%  \\
         \midrule
         \multirow{4}{*}[-1ex]{\centering ASR-30}&  Att-1 &  78.42\% & 73.27\% & 35.49\% \\[1.5pt]
           &  Att-2 &89.34\% &94.39\% &96.41\% \\[1.5pt]
          ~ &  Att-3 & 97.01\% &78.91\% &93.60\% \\[1.5pt]
            \cdashline{2-5} 
            \addlinespace[0.4ex]
          ~ & Avg. &  \textbf{88.26\%} &  82.19\% &  75.17\% \\
         \bottomrule
    \end{tabular*}
    }
\end{table}

Motivated by prior studies on frequency-domain backdoors, we hypothesize that the perturbation location may affect attack performance. As described in Section \ref{5.1}, each attacker perturbs a 3*3 block in the frequency domain. Following the approach of\cite{wang2022invisible}, we position this block along the diagonal of the frequency matrix, with its top-left coordinate denoted as ($f$, $f$),  so the starting position of the frequency block is abbreviated as $f$.  We experimented with different $f$ combinations for the three attackers, and the results on CIFAR-10 are shown in \tabref{tab8}. The combination $f = 15/20/25$ consistently achieves strong attack performance, with all attackers maintaining high ASR and ASR-30 values. The combination $f=15/20/25$ consistently achieves strong attack performance, with all malicious objectives maintaining high ASR and ASR-30 values.

At present, it remains difficult to precisely characterize the influence of $f$ on attack effectiveness or identify a universally optimal combination, since the outcome may also depend on factors such as image content and feature distribution. Nevertheless, some empirical patterns can still be observed. In general, perturbing relatively higher-frequency regions, or enlarging the separation among the selected $f$ values of different malicious objectives, tends to reduce gradient-direction conflicts and improve the stability of multiple backdoors. This observation is broadly consistent with previous findings on frequency-domain backdoor design~\cite{hou2023stealthy},\cite{khayam2003discrete}.

In addition, \tabref{tab9} reports the corresponding stealthiness results for different $f$ values on CIFAR-10. The results show that perturbations in relatively higher-frequency regions usually exhibit better stealthiness. This is because low-frequency components contain the dominant visual information of the image~\cite{xiao2022invisible}, and perturbing them is more likely to introduce noticeable visual artifacts.

\begin{table}[!t]
    \renewcommand{\arraystretch}{1.0} 
    \centering 
    \caption{Effect of frequency block position on stealth performance in CIFAR-10.}
    \label{tab9}
    \setlength{\tabcolsep}{1mm}{
        \begin{tabular*}{\columnwidth}{@{\extracolsep{\fill}}cc @{\hspace{5mm}} ccccc }
            \toprule  
            \multicolumn{2}{c}{Positions}&   5 &  10 &  15 & 20 & 25 \\
            \midrule 
            \multicolumn{2}{c}{ACC}&   72.20\% &  72.32\% &  73.30\% & 73.12\% & 73.57\%\\
            \midrule 
            \multirow{4}{*}[-1.2ex]{\centering SSIM} &  Att-1 &  0.98507 &  0.98454 &  0.98446 &  0.98447 &  0.98435\\[1.5pt]
            &  Att-2 &  0.98460 &  0.98403 &  0.98395 &  0.98396 & 0.98385\\[1.5pt]
            ~ &  Att-3 & 0.98403 &  0.98347 & 0.98338 &  0.98340 &  0.98328\\[1.5pt]
            \cdashline{2-7} 
            \addlinespace[0.4ex]
            ~ & Avg. & 0.98457 & 0.98401 & 0.98393 & 0.98394 & 0.98383\\
            \midrule
            \multirow{4}{*}[-1.2ex]{\centering LPIPS}&  Att-1 &  0.0333 &  0.0257 & 0.0283 &  0.0248 & 0.0067 \\[1.5pt]
            &  Att-2 &  0.0416 & 0.0457&  0.0515&  0.0445 &  0.0107 \\[1.5pt]
            ~ &  Att-3 &  0.0212 &  0.0135 & 0.0125& 0.0170 &  0.0074\\[1.5pt]
            \cdashline{2-7} 
            \addlinespace[0.4ex]
            ~ & Avg. &  0.0320 &  0.0283 &  0.0308 &  0.0288 & \textbf{0.0083} \\
            \bottomrule 
        \end{tabular*}
    }
\end{table}

\begin{table}[!t]
    \centering 
    \caption{Effect of perturbation magnitude on attack and stealth performance in CIFAR-100.}
    \label{tab10}
    \setlength{\tabcolsep}{1mm}{
        \begin{tabular*}{\columnwidth}{@{\extracolsep{\fill}}cc @{\hspace{5mm}} ccccc}
            \toprule  
            \multicolumn{2}{c}{Magnitudes}&   80 &  90 &  100 & 110 & 120\\
            \midrule 
            \multicolumn{2}{c}{ACC}& 52.47\% & 51.97\% &  52.63\% & 52.75\% & 52.26\% \\
            \midrule 
            \multirow{4}{*}[-1.2ex]{\centering ASR}&  Att-1  &  80.88\% &  91.29\% &  91.19\% &  95.67\% &  93.75\%\\[1.5pt]
             &  Att-2  &  91.08\% &  91.27\% & 95.80\% &  90.61\% &  95.21\% \\[1.5pt]  
            ~ &  Att-3 &  23.65\% &  89.02\% &  92.25\% &  93.10\% &  93.96\% \\[1.5pt]
            \cdashline{2-7} 
            \addlinespace[0.4ex]
            ~ & Avg.  &  65.20\% &  90.53\% &  93.08\% &  93.13\% &  \textbf{94.31\%} \\
            \midrule 
            \multirow{4}{*}[-1.2ex]{\centering  ASR-30}&  Att-1  &  65.72\% &  70.26\% &  69.07\% &  80.56\% &  83.95\% \\[1.5pt]          
            &  Att-2 &  75.82\% &  82.59\% &  87.96\% &  82.86\% &  91.90\% \\[1.5pt]           
            ~ &  Att-3  &  19.77\% & 76.28\% &  58.71\% &  84.08\% &  89.11\% \\[1.5pt]
            \cdashline{2-7} 
            \addlinespace[0.4ex]
            ~ & Avg. &  53.77\% &  76.38\% &  71.91\% &  82.50\% &  \textbf{88.32\%} \\[1.5pt]
            \midrule 
            \multirow{4}{*}[-1.2ex]{\centering  SSIM}& Att-1  & 0.98877 & 0.98655 &  0.98428 &  0.98197 & 0.97965 \\[1.5pt]
             &  Att-2  & 0.98872&  0.98649 &  0.98420 &  0.98188 &  0.97954 \\ [1.5pt]
            ~ &  Att-3 & 0.98813 &  0.98580 &  0.98343 &  0.98102 &  0.97859 \\[1.5pt]
            \cdashline{2-7} 
            \addlinespace[0.4ex]
            ~ & Avg.  &\textbf{0.98854} &  0.98628 &  0.98397 &  0.98162 &  0.97926 \\
            \midrule
            \multirow{4}{*}[-1.2ex]{\centering   LPIPS}&  Att-1  &  0.0180 & 0.0222  &  0.0266 & 0.0312 &  0.0360 \\ [1.5pt] 
            &  Att-2 & 0.0346 &  0.0417 & 0.0489 &  0.0563 & 0.0638\\ [1.5pt]  
            ~ &  Att-3 & 0.0075 & 0.0095 &  0.0116 & 0.0139 & 0.0163 \\[1.5pt]
            \cdashline{2-7} 
            \addlinespace[0.4ex]
            ~ & Avg. & \textbf{0.0200} & 0.0245 & 0.0290 & 0.0338 & 0.0387 \\
            \bottomrule  
        \end{tabular*}
    }
\end{table}

\subsubsection{Impact of perturbation magnitude}
\label{5.4.2}
Following~\cite{wang2022invisible}, we further evaluate the effect of perturbation magnitude within each frequency block. As shown in \tabref{tab10}, the results on CIFAR-100 indicate that increasing the perturbation magnitude generally improves attack effectiveness. However, when the perturbation becomes too strong, the trigger pattern may become more conspicuous, thereby increasing the risk of exposure. Therefore, a suitable perturbation magnitude is necessary to balance attack effectiveness and stealthiness.

\subsubsection{Impact of poisoning rate and replay rate}
\label{5.4.3}
We next analyze the influence of the local poisoning rate $r_b$ and the replay rate $r_{br}$ on the performance of DMBA. The results on CIFAR-100 are shown in \tabref{tab11}. In general, increasing these two ratios improves the effectiveness of the attack, since more poisoned samples and replayed samples strengthen malicious optimization. However, excessively large values of $r_b$ and $r_{br}$ may also degrade the performance of the main task and increase the proportion of anomalous samples in local training, making the attack easier to detect. Therefore, moderate values of $r_b$ and $r_{br}$ are important for balancing effectiveness, stealthiness, and optimization stability in coordinated distributed backdoor injection.

\begin{table}[!t]
    \centering 
    \caption{Effect of poisoning ratio ($r_b$) and replay ratio ($r_{br}$) on attack performance in CIFAR-100.}
    \label{tab11}
    \setlength{\tabcolsep}{1mm}{

    \begin{tabular*}{\columnwidth}{@{\extracolsep{\fill}}cc @{\hspace{5mm}} ccccc }
         \toprule  
          \multicolumn{2}{c}{ $r_{b}/ r_{br}$} & 4/1 &  4/2 &  8/3 & 10/4 & 12/5\\
          \midrule
          \multicolumn{2}{c}{ACC}& 52.99\% &  53.19\% &  52.63\% & 51.96\% & 48.31\% \\
         \midrule 
         \multirow{4}{*}[-1.2ex]{\centering ASR}&  Att-1 &  22.02\% &  93.20\% &  91.19\% & 94.62\% &  96.46\% \\[1.5pt]
           &  Att-2 & 85.84\% &  81.06\% &  95.80\% &  94.25\% &  98.12\% \\[1.5pt]
          ~ &  Att-3 &  3.16\% &  61.70\% &  92.25\% &  94.38\% &  96.48\% \\[1.5pt]
          \cdashline{2-7} 
            \addlinespace[0.4ex]
          ~ & Avg. & 37.01\% & 78.65\% & 93.08\% & 94.42\% & \textbf{97.02\%}\\
          \midrule 
          \multirow{4}{*}[-1.2ex]{\centering ASR-30 }&  Att-1 &  17.63\% &  78.28\% &  69.07\% & 85.28\% &  83.22\% \\[1.5pt]
          &  Att-2 & 76.86\% &  52.43\% & 87.96\% &  87.52\% &  92.88\% \\[1.5pt]
          ~ &  Att-3 & 1.38\% &  69.47\% &  58.71\% &  85.16\% &  88.18\% \\[1.5pt]
           \cdashline{2-7} 
            \addlinespace[0.4ex]
         ~ & Avg. & 31.96\% & 66.73\% & 71.91\% & 85.99\% & \textbf{88.09\%} \\
        \bottomrule  
    \end{tabular*}
    }
\end{table}
\begin{table}[!t]
    \centering 
    \caption{Effect of frequency block size on attack performance in CIFAR-100.}
    \label{tab12}
    \setlength{\tabcolsep}{2.5mm}{
        \begin{tabular*}{\columnwidth}{@{\extracolsep{\fill}}cc @{\hspace{12.5mm}} ccc }
         \toprule 
          \multicolumn{2}{c}{\hspace*{-7mm}Block Sizes}& 1*1 & 2*2 & 3*3\\
         \midrule 
         \multicolumn{2}{c}{\hspace*{-7mm}ACC}  &  52.89\% &  52.35\% &  52.63\% \\
         \midrule
         \multirow{4}{*}[-1.2ex]{\centering ASR }&  Att-1  &  3.57\% &  0.33\% &  91.19\% \\[1.5pt]
         &  Att-2  & 34.49\% &  99.1\% & 95.80\%\\[1.5pt]
          ~ &  Att-3 & 2.11\%  & 3.32\% & 92.25\% \\[1.5pt] 
           \cdashline{2-5} 
            \addlinespace[0.4ex]
          &  Avg.  & 13.39\% & 34.25\% & \textbf{93.08\%} \\
         \midrule
          \multirow{4}{*}[-1.2ex]{\centering ASR-30 }&  Att-1 & 0.12\%  & 0.32\% & 69.07\%  \\[1.5pt]
           &  Att-2 & 0.11\% &99.77\% &87.96\%	\\[1.5pt]
          ~ &  Att-3 &0.14\% &2.56\%	&58.71\% \\[1.5pt] 
          \cdashline{2-5} 
            \addlinespace[0.4ex]
          &  Avg.   & 0.12\% & 34.22\% & \textbf{71.91\%} \\
        \bottomrule 
    \end{tabular*}
    }
\end{table}

\subsubsection{Impact of frequency block size}
\label{5.4.4}

We further investigate whether the size of the perturbed frequency block influences the performance of DMBA. To this end, we test block sizes of 1*1, 2*2, 3*3 while keeping the perturbation amplitude unchanged. As shown in \tabref{tab12} using overly small frequency blocks may cause some malicious objectives to fail on CIFAR-100, likely because the corresponding trigger features are too weak to be learned reliably. Increasing the block size strengthens the learned trigger features and improves attack effectiveness. However, excessively large perturbation regions are more likely to introduce visible artifacts. Therefore, the block size should be selected to improve effectiveness while preserving sufficient stealthiness.

\subsubsection{Impact of attack interval}
\label{5.4.5}
Inspired by prior studies on FL backdoor attacks~\cite{xie2019dba},\cite{bagdasaryan2020backdoor}, we further explore whether the interval between attack rounds affects the performance of DMBA. Specifically, we consider fixed intervals of 2, 5, and 10 rounds, as well as a randomly varying interval. The results on GTSRB are summarized in \tabref{tab13}. The results indicate that if attacks are launched too frequently, some malicious objectives may suffer from stronger gradient interference. On the other hand, if the interval is too large, the previously injected backdoors may weaken because of continual benign training and forgetting effects. Therefore, a moderate attack interval is beneficial because it balances malicious-update stability and backdoor persistence over time.

\begin{table}[!t]
    \centering 
    \caption{Effect of attack interval on attack performance in GTSRB.}
    \label{tab13}
    \setlength{\tabcolsep}{1.7mm}{
        \begin{tabular*}{\columnwidth}{@{\extracolsep{\fill}}cc @{\hspace{8.5mm}} cccc}
            \toprule  
            
            \multicolumn{2}{c}{\hspace*{-5mm}Intervals}& 2 &  5 &  10 &  Rnd.\\
            \midrule 
            \multicolumn{2}{c}{\hspace*{-5mm}ACC}  &  98.13\% & 98.36\% &  98.05\% &  99.07\%\\
            \midrule 
            \multirow{4}{*}[-1.2ex]{\centering ASR }&  Att-1 &  98.72\%&  99.61\%&  99.80\%&  99.53\%\\[1.5pt]
             &  Att-2& 99.83\%&  99.56\%&  99.97\%& 99.91\%	\\[1.5pt]
            ~ &  Att-3  &  99.31\% &  98.89\%&  99.97\%&  99.60\%\\[1.5pt]
            \cdashline{2-6} 
            \addlinespace[0.4ex]
            & Avg. & 99.29\% & 99.35\% & \textbf{99.91\%} & 99.68\%\\
            \midrule 
            \multirow{4}{*}[-1.2ex]{\centering ASR-30 }&  Att-1 &	96.79\% &  94.32\%&  98.79\%&  96.54\%\\[1.5pt]
             &  Att-2  &99.69\% &  97.42\%&  100\%&  99.84\%\\[1.5pt]
            ~ &  Att-3 &99.25\%&  99.49\%&  99.20\%&  99.85\%\\[1.5pt]
            \cdashline{2-6} 
            \addlinespace[0.4ex]
            & Avg.& 98.58\% & 97.08\% & \textbf{99.33\%} & 98.74\%\\
            \bottomrule 
        \end{tabular*}
    }
\end{table}

\begin{table}[!t]
    \centering 
    
    \caption{Effect of target label selection on attack performance in CIFAR-10.}
    \label{tab14}
    \setlength{\tabcolsep}{2.5mm}{
            \begin{tabular*}{\columnwidth}{@{\extracolsep{\fill}}cc @{\hspace{12.5mm}} ccc }
                \toprule  
                \multicolumn{2}{c}{\hspace*{-7mm}Target Labels}& [0, 4, 6] & [1, 3, 9] & [2, 7, 8]\\
                \midrule 
                \multicolumn{2}{c}{\hspace*{-7mm}ACC} &  73.30\% &  68.78\% &  70.97\% \\
                \midrule 
                \multirow{4}{*}[-1.2ex]{\centering ASR}&  Att-1 &  96.91\% &  95.80\% &  97.01\% \\[1.5pt]
                &  Att-2 &  94.49\%	&97.29\%	&99.39\%	\\[1.5pt]
                ~ &  Att-3 &  99.28\% &  94.91\% &  95.13\%  \\[1.5pt]
                \cdashline{2-5} 
                \addlinespace[0.4ex]
                &  Avg. & 96.89\% & 96.00\% & 97.18\% \\ 
                \midrule 
                \multirow{4}{*}[-1.2ex]{\centering ASR-30 }&  Att-1 &  78.42\% & 96.49\% & 94.09\% \\[1.5pt]
                &  Att-2 &89.34\% &81.14\% &88.01\% \\[1.5pt]
                ~ &  Att-3 &97.01\%	&96.99\% &85.46\% \\[1.5pt]
                \cdashline{2-5} 
                \addlinespace[0.4ex]
                &  Avg. & 88.26\% & 91.54\% & 89.19\% \\
                \bottomrule 
        \end{tabular*}
    }   
\end{table}
\subsubsection{Impact of target label selection}
\label{5.4.6}
Finally, we investigate whether the selection of target labels affects the performance of DMBA. To this end, we evaluate several combinations of target labels $[t1, t2, t3]$, and the results on CIFAR-10 are reported in \tabref{tab14}. Overall, the choice of target labels does not significantly affect the attack performance in our setting, suggesting that DMBA is relatively robust to different target-label assignments.

\subsection{Extended Evaluations and Discussions}
To further validate the generality and practicality of DMBA, we extend the evaluation from three aspects: broader attack scenarios, runtime overhead, and scalability to more malicious clients. These experiments supplement the main results by examining whether DMBA remains effective beyond the standard targeted setting, whether its additional cost is acceptable in practice, and how its performance changes as the number of malicious clients increases.

\begin{table}[!t]
    \renewcommand{\arraystretch}{0.5} 
    \centering
    \footnotesize
    \caption{DMBA attack performance evaluation in different scenarios.}
    \label{tab15}
    \setlength{\tabcolsep}{0.8mm} 

    \resizebox{\columnwidth}{!}{%
        \setlength{\heavyrulewidth}{0.1mm}  
        \setlength{\lightrulewidth}{0.1mm}  
        \setlength{\arrayrulewidth}{0.05mm}
        \begin{tabular}{ccccccc}
            \toprule
            Attack Scenarios & Datasets & Trigger/Label Selection & Att-1 ASR & Att-2 ASR & Att-3 ASR & ASR Avg. \\
            \midrule
            S1: Targeted-Base & CIFAR-10 & t1→\#0, t2→\#4, t3→\#6 & 96.73\% & 93.97\% & 97.06\% & 95.92\% \\
            S2: Targeted-Label overlap & CIFAR-10 & t1/t2/t3→\#2 & 97.68\% & 90.73\% & 98.45\% & 95.62\% \\
            S3: Targeted-Trigger overlap & CIFAR-10 & t1→\#0/\#4/\#6 & 95.40\% & 0.00\% & 0.00\% & 31.80\% \\
            S4: Non-targeted & CIFAR-10 & t1/t2/t3→\#random & 96.74\% & 84.29\% & 92.65\% & 91.22\% \\
            S5: Non-targeted & CIFAR-100 & t1/t2/t3→\#random & 98.80\% & 84.63\% & 98.14\% & 93.86\% \\
            S6: Mixed (1Targeted+2Non-targeted) & CIFAR-10 & t1→\#2, t2/t3→\#random & 97.19\% & 86.01\% & 87.86\% & 90.35\% \\
            S7: Mixed (2Targeted+1Non-targeted) & CIFAR-10 & t1→\#2, t2→\#4, t3→\#random & 96.80\% & 78.55\% & 92.77\% & 89.37\% \\
            \bottomrule
        \end{tabular}%
    }
\end{table}

\begin{table}[!t]
    \renewcommand{\arraystretch}{0.5} 
    \centering
    \footnotesize
    \caption{Runtime Results over 100 Epochs.}
    \label{tab16}
    \setlength{\tabcolsep}{0.8mm} 

    \resizebox{\columnwidth}{!}{%
        \setlength{\heavyrulewidth}{0.1mm}  
        \setlength{\lightrulewidth}{0.1mm}  
        \setlength{\arrayrulewidth}{0.05mm}
        \begin{tabular}{cccccc}
            \toprule
            Configuration & Run 1 (s) & Run 2 (s) & Run 3 (s) & Avg. Time (s) & Relative to Baseline \\
            \midrule
            Baseline (CFCT+BR) & 2160.67 & 2250.23 & 2277.12 & 2229.34 & 100\% \\
            Without CFCT & 1968.89 & 2058.45 & 2085.34 & 2037.56 & 91.4\% \\
            Without BR & 2160.59 & 2250.15 & 2277.04 & 2229.26 & 99.996\% \\
            Without CFCT+BR & 1968.81 & 2058.37 & 2085.26 & 2037.48 & 91.4\% \\
            \bottomrule
        \end{tabular}
    }
\end{table}

\subsubsection{Evaluation under different attack scenarios}

Although the main design of DMBA is presented under the targeted setting, its underlying motivation is more general: the core difficulty lies in mitigating the interference among multiple malicious objectives injected through distributed malicious clients. To examine this point, we evaluate DMBA under several different scenarios, including targeted, non-targeted, and mixed settings. The results are summarized in \tabref{tab15}.

In the standard targeted setting (S1), DMBA achieves strong and balanced attack performance, with all three malicious objectives maintaining high ASRs and an average ASR of 95.92\%. When target labels partially overlap (S2), the average ASR remains similarly high at 95.62\%, indicating that moderate overlap in target selection does not substantially weaken the attack. However, when trigger-target assignments are made highly similar (S3), the performance becomes unstable: one malicious objective almost completely fails, and the average ASR drops to 31.80\%. This result suggests that excessive overlap among malicious objectives can intensify interference and significantly degrade coordinated multi-backdoor injection.

We further evaluate non-targeted scenarios. In S4, where all three malicious objectives are non-targeted, DMBA still achieves an average ASR of 91.22\%, showing that the method is not restricted to purely targeted attacks. In S5, where all three backdoors are non-targeted but share the same randomly assigned target behavior, the average ASR further increases to 93.86\%, suggesting that DMBA remains effective when malicious objectives are less explicitly constrained by fixed target labels. These results support the view that the proposed method addresses a broader optimization problem than targeted attacks alone.

Finally, we consider mixed settings that combine targeted and non-targeted malicious objectives. As shown in \tabref{tab15}, both mixed scenarios maintain strong attack performance, with average ASRs of 90.35\% in S6 and 89.37\% in S7. Although these values are slightly lower than those of the standard targeted setting, they remain consistently high across different objective combinations. Overall, these results indicate that DMBA can generalize beyond the base targeted setting and remains effective in more diverse attack scenarios, provided that the overlap among malicious objectives is not excessively strong. This observation is consistent with the design rationale of DMBA, which aims to mitigate gradient conflicts rather than optimize only one specific target-assignment pattern.

\subsubsection{Runtime overhead over 100 epochs}

In addition to attack effectiveness, practical overhead is also an important concern, especially in resource-constrained FL environments. To complement the complexity analysis in Section \ref{4.4}, we measure the runtime of different configurations over 100 epochs. The results are reported in \tabref{tab16}. 

The full DMBA configuration with both CFCT and BR requires an average runtime of 2229.34 seconds and is used as the baseline reference. Removing CFCT reduces the average runtime to 2037.56 seconds, corresponding to 91.4\% of the baseline runtime. By contrast, removing BR has almost no effect on runtime, with an average runtime of 2229.26 seconds, i.e., 99.996\% of the baseline. When both CFCT and BR are removed, the average runtime is 2037.48 seconds, again corresponding to 91.4\% of the baseline.

\begin{figure}[!t]
    \centering 
    \includegraphics[width=1\linewidth]{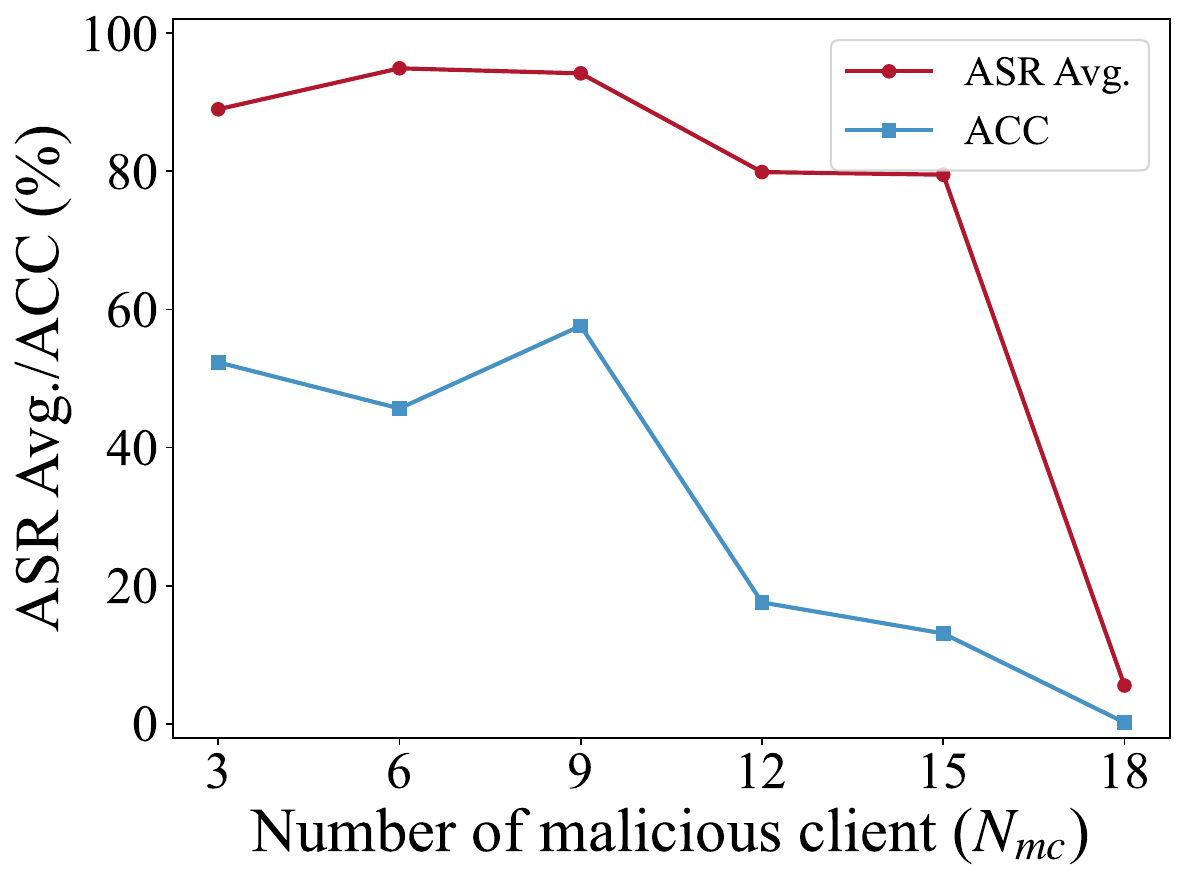} 
    \caption{Backdoor and main task performance with different numbers of malicious clients (backdoors).} 
    \label{fig6}
\end{figure}

These results show that the additional runtime introduced by DMBA mainly comes from CFCT, whereas the BR mechanism contributes negligible runtime overhead. More importantly, even the full DMBA configuration increases runtime by only about 8.6\% relative to the setting without CFCT and BR, which is consistent with the lightweight complexity analysis in Section \ref{4.4}. Therefore, the proposed method preserves practical feasibility: BR introduces almost no additional runtime burden, and the overhead of CFCT remains moderate and predictable.

\subsubsection{Scalability to more malicious clients}

We further investigate how DMBA behaves when the number of malicious clients increases. The corresponding results are shown in \figref{fig6}, where both the average ASR and the clean-task ACC are plotted against the number of malicious clients $N_{mc}$.

When the number of malicious clients increases from 3 to 6 and then to 9, DMBA maintains strong attack performance, with the average ASR remaining close to or above 90\%. This indicates that the proposed design can effectively support a moderate increase in the number of malicious objectives while preserving joint backdoor effectiveness. However, as $N_{mc}$ urther increases to 12 and 15, the average ASR begins to decline to around 80\%, and the clean accuracy also drops substantially. When $N_{mc}=18$, the attack performance collapses, and the model utility is almost completely lost. 

This trend reveals an important scalability characteristic of DMBA. On the one hand, the proposed combination of BR and CFCT is effective for coordinated distributed attacks involving a moderate number of malicious clients. On the other hand, the benefit of increasing the number of malicious clients is not unlimited. As the number of malicious objectives grows, the overlap and competition among them become increasingly difficult to control, and the model capacity available to preserve multiple distinguishable backdoors becomes more constrained. In such cases, the conflict-mitigation effect of CFCT and BR exhibits diminishing returns, and the accumulated optimization burden eventually overwhelms the model’s ability to maintain both clean performance and multiple backdoor behaviors.

\subsubsection{Discussion of limitations}
The above extended evaluations also reveal several limitations of the current DMBA design. First, although DMBA generalizes well to targeted, non-targeted, and mixed settings, its performance can deteriorate when the overlap among malicious objectives becomes too strong, as shown in scenario S3. This indicates that the method still depends on maintaining a sufficient degree of distinguishability among malicious objectives.

Second, while DMBA scales effectively to a moderate number of malicious clients, its performance degrades when too many malicious objectives are injected simultaneously. This limitation is not merely due to implementation details, but is also related to the finite representational capacity of the model and the increasing optimization conflicts among multiple backdoor objectives. In particular, the advantage brought by CFCT becomes marginal when the number of malicious clients grows too large, because channel-frequency separation alone cannot fully eliminate interference under heavily crowded malicious objectives.

Third, although the measured runtime overhead remains moderate, the current experiments are still conducted on relatively small image benchmarks and standard backbone models. Therefore, the computational and optimization behavior of DMBA under larger image resolutions, more complex architectures, and more realistic large-scale FL deployments remains to be studied further.

These observations do not weaken the main conclusion of this work, but rather clarify the applicability boundary of DMBA: it is effective and practical for coordinated distributed attacks with a moderate number of malicious clients and sufficiently distinguishable malicious objectives, while more challenging large-scale settings require further study. We will revisit these limitations in the Conclusion and discuss corresponding future directions.


\subsection{Robustness of DMBA against defenses}
\label{5.6}
\begin{figure*}[!t]
\centering
\subfloat[ClippedClustering\label{fig7a}]{
    \includegraphics[width=0.3\linewidth]{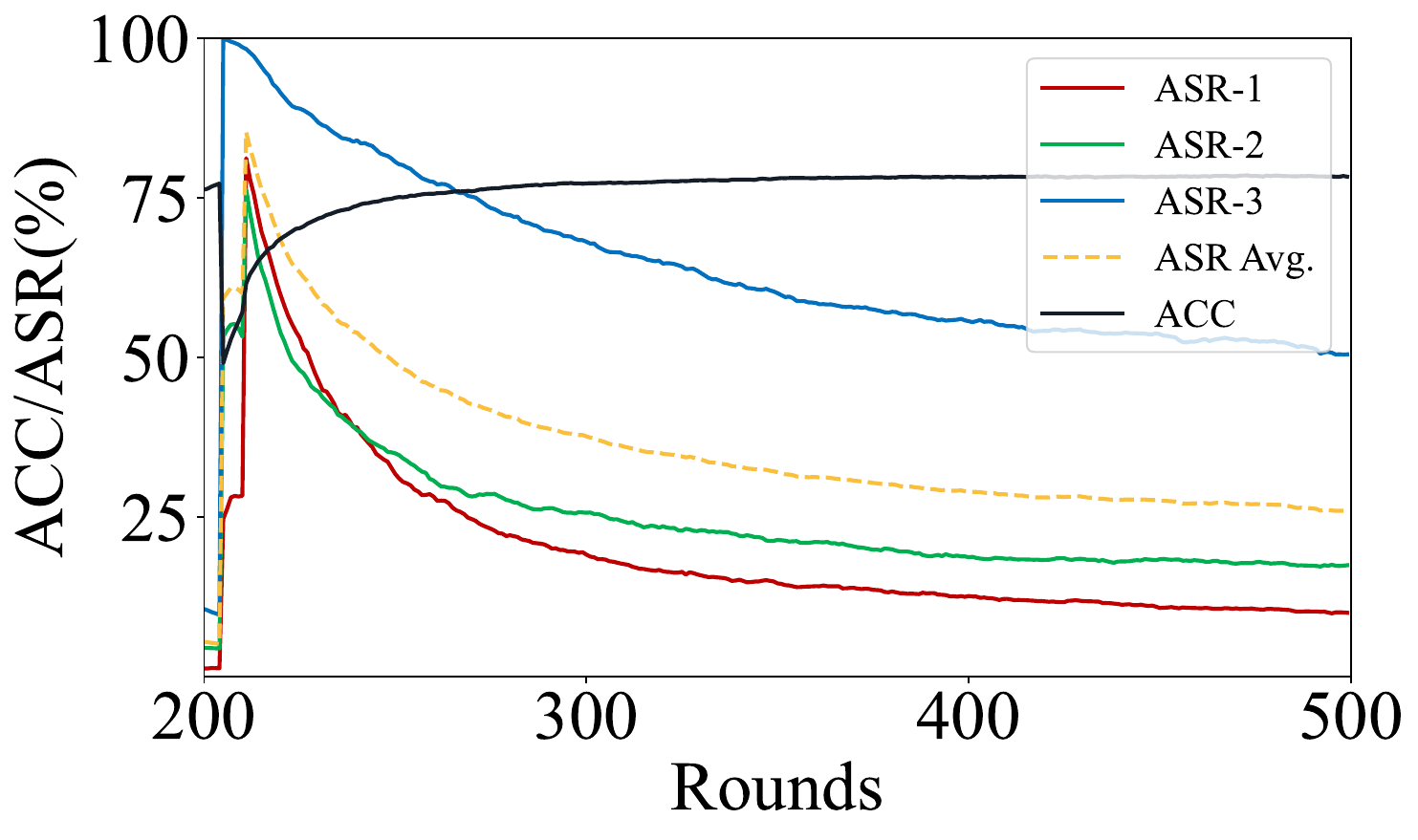}
}
\hfill
\subfloat[DP-FedAvg\label{fig7b}]{
    \includegraphics[width=0.3\linewidth]{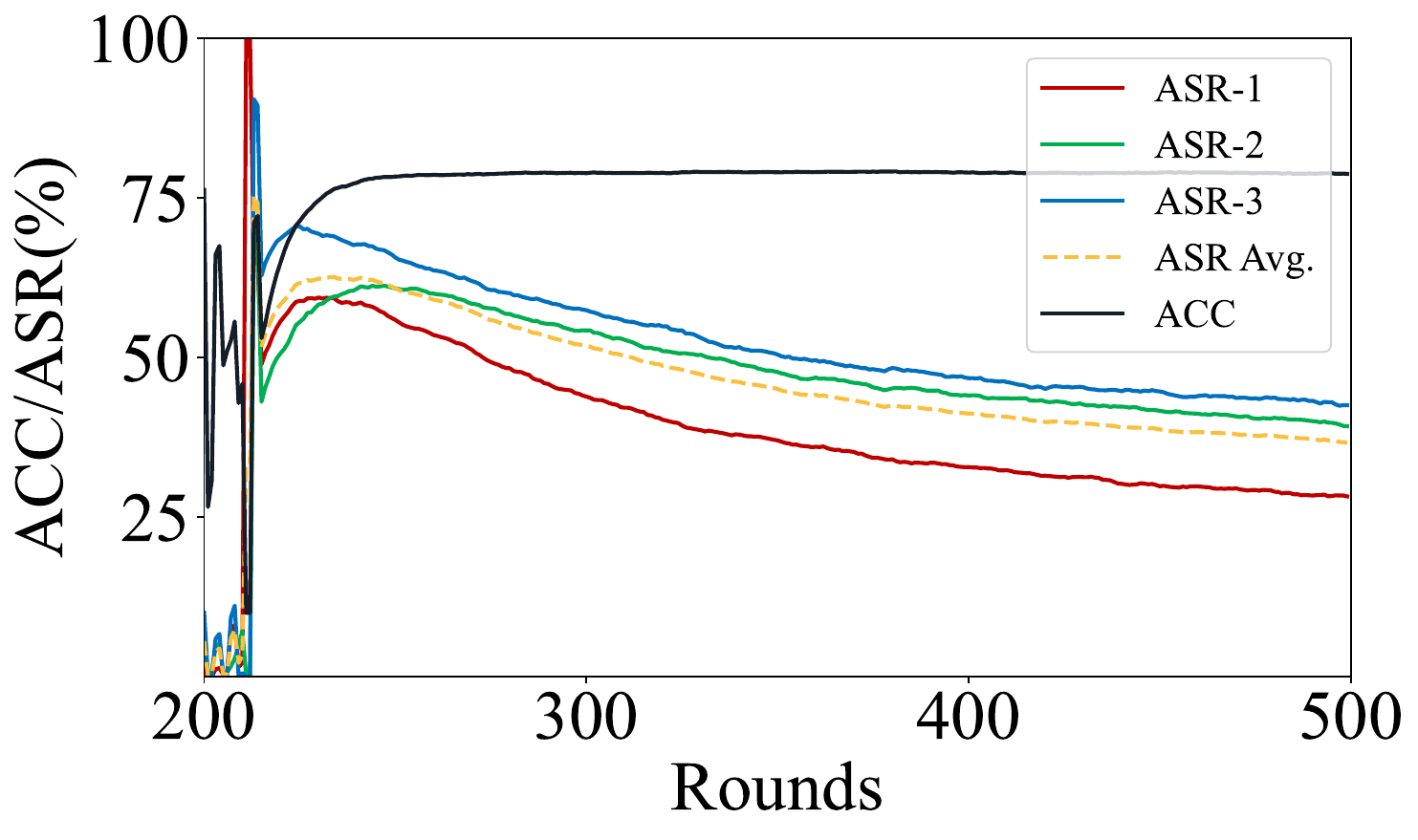}
}
\hfill
\subfloat[SignGuard\label{fig7c}]{
    \includegraphics[width=0.3\linewidth]{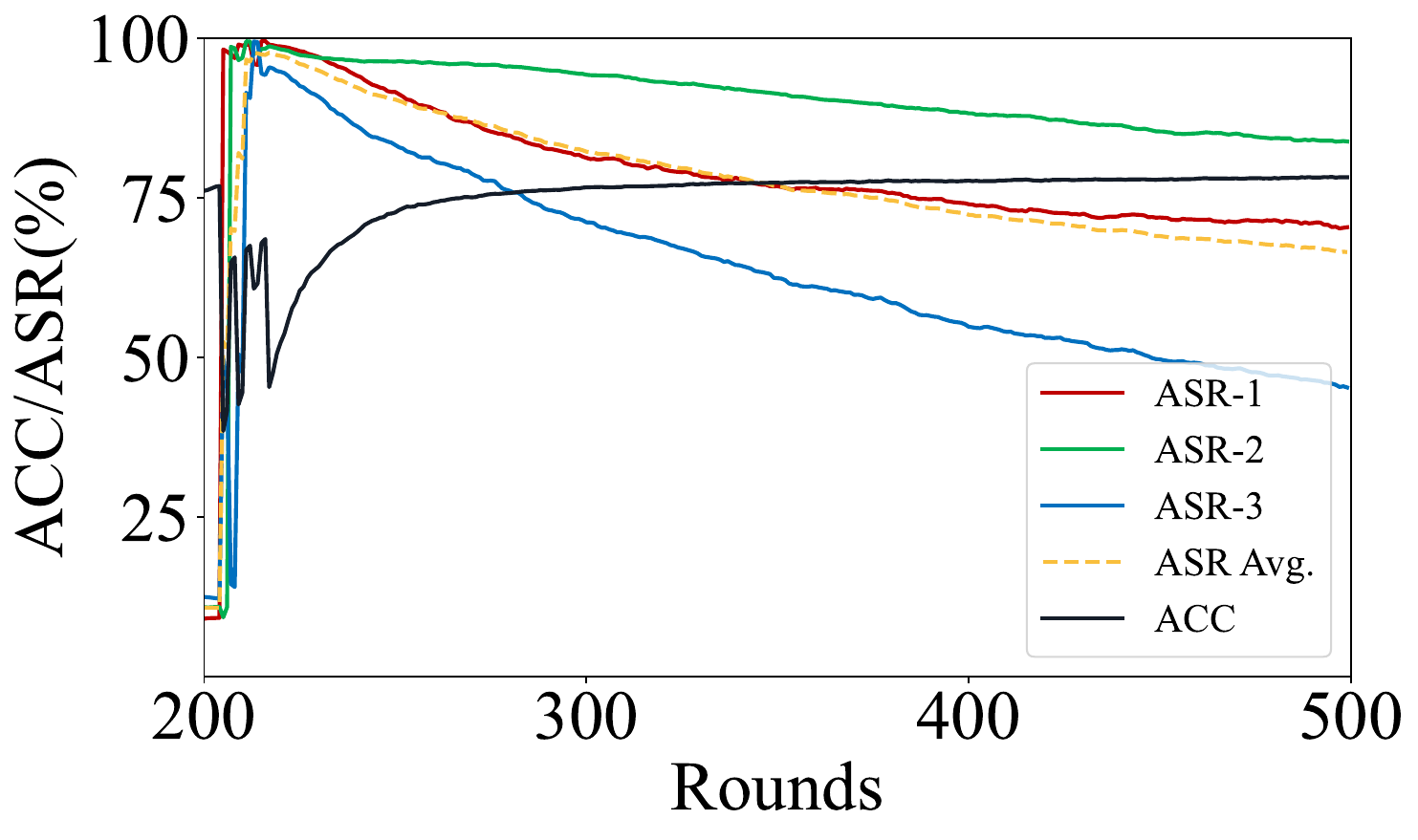}
}
\caption{Performance of DMBA on different defense methods on CIFAR-10.}
\label{fig7}
\end{figure*}

This section evaluates the robustness of DMBA against three representative defenses introduced in Section \ref{2.2}: ClippedClustering, DP-FedAvg, and SignGuard, which correspond to robust aggregation, differential-privacy-based protection, and malicious-gradient filtering, respectively. \figref{fig7} presents the ASR trajectories of different malicious objectives on CIFAR-10 under these defenses. Overall, DMBA remains effective across all settings, although the degree of degradation varies. Notably, several malicious objectives still maintain relatively high ASR values even after the attack phase ends, indicating that existing defenses cannot fully remove the implanted backdoor behaviors.

The results suggest that the resilience of DMBA stems from its coordinated multi-objective design. The BR mechanism reduces discrepancies among malicious updates before aggregation, making them less distinguishable from benign gradients, while CFCT improves trigger separability while maintaining visual stealthiness. Consequently, although the defenses may weaken some objectives, they fail to consistently suppress all backdoors in the distributed multi-target scenario. This finding indicates that robustness against conventional single-backdoor attacks does not necessarily translate to robustness in coordinated multi-objective settings, highlighting the need for defense strategies specifically designed for such attacks.

\section{Conclusion}
In this paper, we proposed DMBA, a coordinated distributed multi-target backdoor attack for federated learning under a practical threat model where a single adversarial entity controls multiple malicious clients with distinct backdoor objectives. Unlike conventional single-backdoor or centralized multi-target settings, this formulation captures a more realistic distributed attack scenario in IoT-oriented FL systems. To mitigate interference among multiple malicious objectives, DMBA integrates two components: the Backdoor Replay (BR) mechanism, which reduces conflicts among malicious updates during aggregation, and the Channel-Frequency Composite Trigger (CFCT) strategy, which enhances trigger distinguishability and alleviates local interference. Experimental results demonstrate that DMBA can effectively preserve multiple distinguishable backdoors while maintaining stealthiness and remaining resilient to representative defenses.

Despite these advantages, DMBA has certain limitations. Its effectiveness may decrease when objective overlap becomes substantial or when the number of malicious clients grows excessively. Future work will investigate larger-scale settings, stronger model architectures, and broader FL applications, as well as potential defenses against coordinated multi-objective backdoor attacks.


{\appendix[Additional Hyperparameter Analysis Across Datasets]
Here provides a comprehensive supplementary hyperparameter analysis on additional datasets. The evaluation is conducted from both attack effectiveness and stealthiness perspectives, and the results consistently support the conclusions drawn in the main text.
\subsection{Effect of Frequency Block Position}
\label{appendixA}
This section examines the impact of the initial position of the frequency blocks on the attack and stealth performance of DMBA on CIFAR-100 and GTSRB. Table \ref{Table A.1} illustrates that DMBA achieves optimal attack efficiency and persistence with the initial positions of frequency blocks set at ``15/20/25''. Additionally, Table \ref{Table A.2} indicates that targeting the high-frequency region for perturbation enhances stealthiness. This is because the low-frequency region carries the primary information of the image, and perturbing it is more likely to result in noticeable visual anomalies.The above conclusions are consistent with the main text.

\renewcommand{\thetable}{A.\arabic{table}}
\setcounter{table}{0}   

\begin{table}[H]
    \renewcommand{\arraystretch}{1.3} 
    \centering 
    \tiny 
    \vspace{-2mm}
    \caption{Effect of frequency block position on attack performance.}
    \label{Table A.1}
    \setlength{\tabcolsep}{2mm}{
        \begin{tabular}{cc c c c@{\hspace{5.6mm}} c c c}
         \toprule [0.4mm] 
          \multicolumn{2}{c}{Datasets} & \multicolumn{3}{c}{CIFAR-100} & \multicolumn{3}{c}{GTSRB}\\
         \midrule  
          \multicolumn{2}{c}{Frequency Block Position}& 15/20/25 & 15/15/15 & 5/15/25 & 15/20/25 &  15/15/15 &  5/15/25\\
         \midrule  
         \multicolumn{2}{c}{ACC} & 52.63\% & 50.33\% & 52.38\% & 98.13\% & 97.05\% & 97.72\%\\
         \midrule  
         \multirow{3}*{~} &  Att-1 & 91.19\% & 89.58\% & 8.41\% & 98.72\% & 99.92\% & 90.18\%\\
         \cmidrule{3-8}  
          ASR &  Att-2 &95.80\% & 96.34\% & 94.16\% & 99.83\% &99.36\%	& 99.66\%\\
         \cmidrule{3-8}  
          &  Att-3 &  92.25\% & 94.48\% & 93.63\% & 99.31\% &  88.58\% & 15.20\% \\
         \midrule  
         \multirow{3}*{~} & Att-1 & 69.07\% & 76.59\% & 0.39\% & 96.79\% &98.63\% & 83.18\% \\
         \cmidrule{3-8}  
          ASR-30 &  Att-2 &87.96\% &81.34\% &75.33\% &99.67\% &98.99\% &99.48\% \\
         \cmidrule{3-8}  
          ~ &  Att-3 &58.71\% &64.41\%	&79.73\% &99.25\%  &60.82\% &0.86\%\\
        \bottomrule [0.4mm] 
    \end{tabular}
    }
    
\end{table}

\begin{table}[H]
    \renewcommand{\arraystretch}{0.5} 
    \centering 
    \tiny 
    \vspace{-2mm}
    \caption{Effect of frequency block position on stealth performance.}
    \label{Table A.2}
    \setlength{\tabcolsep}{1.mm}{
        \begin{tabular}{cc@{\hspace{1.8mm}} c@{\hspace{1.8mm}} c@{\hspace{1.8mm}} c@{\hspace{1.8mm}} c@{\hspace{1.8mm}} c@{\hspace{3.6mm}} c@{\hspace{1.8mm}} c@{\hspace{1.8mm}} c@{\hspace{1.8mm}} c@{\hspace{1.8mm}} c}
         \toprule [0.4mm] 
          \multicolumn{2}{c}{Datasets} &   \multicolumn{5}{c}{CIFAR-100} &  \multicolumn{5}{c}{GTSRB}\\
         \midrule  
          \multicolumn{2}{c}{Location}& 5 &  10 &  15 & 20 & 25 & 5 &  10 & 15 & 20 & 25\\
         \midrule  
         \multirow{3}*{~} & Att-1 &  0.98487 & 0.98436 &  0.98428 &  0.98429 & 0.98418 &  0.97780 &  0.97708 &  0.97694 &  0.97697 &  0.97683   \\
         \cmidrule{3-12}  
          SSIM &  Att-2 &  0.98483 &  0.98428 &  0.98420 &  0.98422 &  0.98410 &  0.97772 & 0.97711 &  0.97697 &  0.97700 &  0.97685\\
         \cmidrule{3-12}  
          ~ &  Att-3 & 0.98403 &  0.98351 &  0.98343 &  0.98344 &  0.98333 &  0.97874 & 0.97805 & 0.97792 & 0.97795 &  0.97780\\
    
          \midrule  
         \multirow{3}*{~}  & Att-1 &  0.0309 &  0.0240 &  0.0266 & 0.0234 &  0.0062 &  0.05020 & 0.04606 &  0.05052 & 0.04630 &  0.01487\\
         \cmidrule{3-12}  
          LPIPS &  Att-2 &  0.0380 &  0.0431& 0.0489 &  0.0424 & 0.0099 &  0.06985 &  0.07629 & 0.08314 & 0.07518 &  0.02243\\
         \cmidrule{3-12}  
          ~ &  Att-3 & 0.0194 & 0.0122&  0.0116 & 0.0159 & 0.0069 &  0.03342 & 0.02569&  0.02482 &  0.03365&  0.01630\\
        \bottomrule [0.4mm] 
    \end{tabular}
    }
    
\end{table}
\subsection{Effect of Perturbation Magnitude}
\label{appendixF2}
This section analyzes the impact of perturbation magnitude on the attack and stealth performance of backdoor attacks on CIFAR-10 and GTSRB. According to Table \ref{Table B.1} and Table \ref{Table B.2}, increasing the perturbation amplitude enhances the ASR and persistence of the backdoor. However, this comes at the cost of reduced invisibility. Thus, selecting the appropriate perturbation amplitude requires careful consideration to balance effectiveness and stealth.The above conclusions are consistent with the main text.

\renewcommand{\thetable}{B.\arabic{table}}
\setcounter{table}{0}   

\begin{table}[H]
    \renewcommand{\arraystretch}{0.5} 
    \centering 
    \tiny 
    \vspace{-2mm}
    \caption{Effect of perturbation magnitude on attack performance.}
    \label{Table B.1}
    \setlength{\tabcolsep}{1.mm}{
        \begin{tabular}{c@{\hspace{1.8mm}}c@{\hspace{1.8mm}}c@{\hspace{1.8mm}}c@{\hspace{1.8mm}}c@{\hspace{1.8mm}}c@{\hspace{1.8mm}}c@{\hspace{3.6mm}} c@{\hspace{1.8mm}}c@{\hspace{1.8mm}}c@{\hspace{1.8mm}}c@{\hspace{1.8mm}}c}
         \toprule [0.4mm] 
          \multicolumn{2}{c}{Datasets} &  \multicolumn{5}{c}{CIFAR-10} &  \multicolumn{5}{c}{GTSRB}\\
         \midrule  
          \multicolumn{2}{c}{\makecell{Perturbation \\ Magnitude}}&   80 &  90 &  100 & 110 & 120 & 80 &  90 & 100 & 110 & 120\\
          \midrule  
          \multicolumn{2}{c}{ACC}&   72.20\% &  72.32\% &  73.30\% & 73.12\% & 73.57\% & 97.14\% &  97.84\% & 98.13\% & 98.05\% & 98.17\%\\
         \midrule  
         \multirow{3}*{~} &  Att-1 &  94.23\% &  98.34\% &  96.91\% &  98.31\% &  89.44\% &  97.76\% &  95.47\% &  98.72\% &  97.81\% &  94.21\%\\
         \cmidrule{3-12}  
          ASR &  Att-2 & 96.60\% & 94.99\% &  94.49\% &  95.90\% &  92.38\% &  99.79\% &  99.95\% &  99.83\% &  99.75\% &  99.85\%\\
         \cmidrule{3-12}  
          ~ &  Att-3 &  98.63\% &  98.80\% & 99.28\% &  99.52\% & 98.36\% &  99.84\% &  99.44\% &  99.31\% &  99.44\% & 99.92\%\\
          \midrule  
         \multirow{3}*{~} &  Att-1 &  73.53\% &  75.38\% &  78.42\% & 86.94\% &  63.80\% &  89.13\% &  82.89\% &  96.79\% &  97.31\% &  94.52\%\\
         \cmidrule{3-12}  
          ASR-30 &  Att-2 & 89.08\% & 87.39\% & 89.34\% &  81.36\% &  92.02\% &  85.50\% &  99.53\% &  99.67\% &  98.78\% &  97.69\%\\
         \cmidrule{3-12}  
          ~ &  Att-3 &  95.33\% & 91.80\% &  97.01\% &  93.86\% &  97.42\% & 99.91\% &  98.91\% &  99.25\% &  98.93\% &  99.83\%\\
        \bottomrule [0.4mm] 
    \end{tabular}
    }
    
\end{table}

\begin{table}[H]
    \renewcommand{\arraystretch}{0.5} 
    \centering 
    \tiny 
    \vspace{-2mm}
    \caption{Effect of perturbation magnitude on stealth performance.}
    \label{Table B.2}
    \setlength{\tabcolsep}{1.mm}{
        \begin{tabular}{c@{\hspace{1.8mm}}c@{\hspace{1.8mm}}c@{\hspace{1.8mm}}c@{\hspace{1.8mm}}c@{\hspace{1.8mm}}c@{\hspace{1.8mm}}c@{\hspace{3.6mm}} c@{\hspace{1.8mm}}c@{\hspace{1.8mm}}c@{\hspace{1.8mm}}c@{\hspace{1.8mm}}c}
         \toprule [0.4mm] 
          \multicolumn{2}{c}{Datasets} &  \multicolumn{5}{c}{CIFAR-10} &  \multicolumn{5}{c}{GTSRB}\\
         \midrule  
          \multicolumn{2}{c}{\makecell{Perturbation \\ Magnitude}}& 80 &  90 &  100 & 110 & 120 & 80 &  90 &  100 & 110 & 120 \\
         \midrule  
         \multirow{3}*{~} & Att-1 &  0.98886 &  0.98668 & 0.98446&  0.98221 &  0.97994 & 0.98313 & 0.98005 &  0.97694 &  0.97384 &  0.97076   \\
         \cmidrule{3-12}  
          SSIM &  Att-2 &  0.98847 &  0.98623 &  0.98395 &  0.98164 & 0.97932 &  0.98308 & 0.98003 &  0.97697 &  0.97393 &  0.97091\\
         \cmidrule{3-12}  
          ~ &  Att-3 & 0.98802 &  0.98572 & 0.98338 &  0.98101 &  0.97864 &  0.98378 & 0.98086 & 0.97792 & 0.97500&  0.97209\\
          \midrule  
        
         \multirow{3}*{~} &  Att-1 &  0.0193 & 0.0237 & 0.0283 &  0.0331 & 0.0381 &  0.0363 & 0.0434 &  0.0505 & 0.0577 & 0.0649\\
         \cmidrule{3-12}  
          LPIPS &  Att-2 &  0.0366 & 0.0439&  0.0515&  0.0592 &  0.0670 &  0.0624 &  0.0728 & 0.0831 & 0.0932 &  0.1032\\
         \cmidrule{3-12}  
          ~ &  Att-3 &  0.0081 &  0.0102 & 0.0125& 0.0150 &  0.0175 &  0.0169 & 0.0208 &  0.02482 &  0.0290 &  0.0333\\
        \bottomrule [0.4mm] 
    \end{tabular}
    }
    
\end{table}
\subsection{Effect of Poisoning Ratio and Backdoor Replay Ratio}
\label{appendixF3}
This section explores the influence of the poisoning ratio and backdoor replay ratio on the performance of backdoor attacks on CIFAR-10 and GTSRB. Table \ref{Table C.1:} demonstrates that higher poisoning ratios and backdoor replay ratios significantly improve the effectiveness of the corresponding backdoors. However, in practical deployments, an excessively high poisoning ratio may increase the risk of detection, thereby compromising the effectiveness and sustainability of the attack. The above conclusions are consistent with the main text.

\renewcommand{\thetable}{C.\arabic{table}}
\setcounter{table}{0}   

\begin{table}[H]
    \renewcommand{\arraystretch}{0.5} 
    \centering 
    \tiny 
    \vspace{-2mm}
    \caption{Impact of poisoning ratio and backdoor replay ratio on attack performance.}
    \label{Table C.1:}
    \setlength{\tabcolsep}{1.mm}{
        \begin{tabular}{cc ccccc@{\hspace{3.6mm}} ccccc }
         \toprule [0.4mm] 
          \multicolumn{2}{c}{Datasets} &  \multicolumn{5}{c}{CIFAR-10} &  \multicolumn{5}{c}{GTSRB}\\
         \midrule  
          \multicolumn{2}{c}{Ratio}&   1/1/4 &  2/2/4 &  3/3/8 & 4/4/10 & 5/5/12 &1/1/4 &  2/2/4 &  3/3/8 & 4/4/10 & 5/5/12 \\
          \midrule  
          \multicolumn{2}{c}{ACC}&  74.31\% &  72.77\% &  73.30\% & 73.02\% & 73.42\% & 98.75\% &  98.06\% & 98.13\% & 97.23\% & 96.76\%\\
         \midrule  
         \multirow{3}*{~} &  Att-1 &  97.04\% &  91.80\% &  96.91\% &  96.71\% &  98.69\% &  91.39\% &  77.93\% & 98.72\% &  98.55\% &  99.30\%\\
         \cmidrule{3-12}  
          ASR &  Att-2 &  92.83\% & 91.29\% &  94.49\% &  96.70\% &  95.16\% &  99.35\% &  97.49\% &  99.83 \%&  99.74\% &  99.95\% \\
         \cmidrule{3-12}  
          ~ &  Att-3 &  13.74\% &  95.72\% &  99.28\% &  99.91\% &  99.18\% &  91.17\% &  7.041\% &  99.31\% &  95.25\% &  100\% \\
          \midrule  
         \multirow{3}*{~} &  Att-1 &  80.03\% &  61.54\% &  78.42\% & 91.09\% &  91.60\% &  78.15\% & 79.12\% &  96.79\% &  81.19\% &  99.02\% \\
         \cmidrule{3-12}  
          ASR-30 &  Att-2 &  91.41\% &  77.59\% &  89.34\% &  83.77\% &  85.76\% &  98.58\% &  88.85\% &  99.67\% &  96.85\% &  99.87\% \\
         \cmidrule{3-12}  
          ~ &  Att-3 &  6.01\%\% & 86.89\% & 97.01\% & 96.74\% &  93.32\% &  70.39\% &  0.82\% &  99.25\% &  79.27\% &  99.87\% \\
        \bottomrule [0.4mm] 
    \end{tabular}
    }
    
\end{table}
\subsection{Effect of Frequency Block Size}
\label{appendixF4}

We hypothesized that the size of the frequency block might influence the attack performance of DMBA. To test this, we experimented with frequency blocks sized 1*1, 2*2, and 3*3 for perturbation. The experimental results, detailed in Table \ref{Table D.1:}, indicate that perturbing larger frequency blocks generally enhances the attack performance of DMBA. However, an excessively large perturbation region may compromise visual stealthiness and reduce the imperceptibility of the attack. The above conclusions are consistent with the main text.

\renewcommand{\thetable}{D.\arabic{table}}
\setcounter{table}{0}   

\begin{table}[H]
    \renewcommand{\arraystretch}{0.5} 
    \centering 
    \tiny 
    \vspace{-2mm}
    \caption{Effect of frequency block size on attack performance.}
    \label{Table D.1:}
    \setlength{\tabcolsep}{1.mm}{
        \begin{tabular}{ccccc@{\hspace{3.6mm}}ccc}
         \toprule [0.4mm] 
          \multicolumn{2}{c}{Datasets} &  \multicolumn{3}{c}{CIFAR-10} &  \multicolumn{3}{c}{GTSRB}\\
         \midrule  
          \multicolumn{2}{c}{Block Size}& 1*1 & 2*2 & 3*3& 1*1 & 2*2 & 3*3\\
         \midrule  
         \multicolumn{2}{c}{ACC} &  73.27\% &  72.79\% &  73.30\% & 98.33\% &  99.10\% &  98.13\%\\
         \midrule  
         \multirow{3}*{~} &  Att-1 &  91.49\% &  87.67\% &  96.91\% &  99.66\% &  96.09\% &  98.72\%\\
         \cmidrule{3-8}  
          ASR &  Att-2 & 97.82\% &  99.03\% &  94.49\% &  99.69\% & 99.86\% &  99.82\% \\
         \cmidrule{3-8}  
          ~ &  Att-3 &  98.88\% & 96.43\% & 99.28\% & 97.22\% &99.88\% &	99.31\% \\
         \midrule  
         \multirow{3}*{~} &  Att-1 &  25.71\% & 92.22\% & 78.42\% & 98.01\% & 94.33\% &	96.79\% \\
         \cmidrule{3-8}  
          ASR-30 &  Att-2 &30.68\% &77.30\% &89.34\% 	 
                    &99.85\% &83.55\% &99.67\% \\
         \cmidrule{3-8}  
          ~ &  Att-3 &36.46\%	&74.51\% &97.01\% &85.90\% &94.30\% &99.25\%\\
        \bottomrule [0.4mm] 
    \end{tabular}
    }
    
\end{table}
\subsection{Effect of Poison Injection Round Interval}
\label{appendixF5}

We hypothesized that the interval between poison injection rounds could impact the performance of DMBA. To investigate, we set the poison injection intervals at fixed rounds of 2, 5, 10, as well as at randomly varying intervals. The experimental outcomes on CIFAR-10 and CIFAR-100, presented in Table \ref{Table E.1}, suggest that extending the interval between rounds tends to enhance DMBA's attack performance. This improvement likely occurs because shorter intervals can cause significant gradient conflicts among different backdoors, undermining the effectiveness of some backdoors.The above conclusions are consistent with the main text.

\renewcommand{\thetable}{E.\arabic{table}}
\setcounter{table}{0}   

\begin{table}[H]
    \renewcommand{\arraystretch}{0.5} 
    \centering 
    \tiny 
    \vspace{-2mm}
    \caption{Effect of poison injection round interval on attack performance.}
    \label{Table E.1}
    \setlength{\tabcolsep}{1.mm}{
        \begin{tabular}{cc cccc@{\hspace{3.6mm}} cccc }
         \toprule [0.4mm] 
          \multicolumn{2}{c}{Datasets} &  \multicolumn{4}{c}{CIFAR-10} & \multicolumn{4}{c}{CIFAR-100} \\
         \midrule  
          \multicolumn{2}{c}{Injection Interval}& 2 &  5 &  10 &  Rondom& 2 &  5 &  10&  Rondom\\
         \midrule  
         \multicolumn{2}{c}{ACC} &  73.30\% &  74.87\% &  75.58\% &  72.20\% &  52.63\% &  53.72\% &  56.83\% &  53.35\%\\
         \midrule  
         \multirow{3}*{~} &  Att-1 &  96.91\% &  97.91\% &  99.90\% &  98.72\% & 91.19\% &  93.20\% &  98.32\% &  94.62\%\\
         \cmidrule{3-10}  
          ASR &  Att-2 &  94.49\%	&99.01\%	&98.16\%	&99.09\%	&95.80\% &90.60\%	&97.48\%	&96.41\%\\
         \cmidrule{3-10}  
          ~ &  Att-3 &  99.28\% &  99.58\% &  99.89\% &  99.81\% &  92.25\% &  3.44\% &  89.92\% &  95.32\% \\
         \midrule  
         \multirow{3}*{~} &  Att-1 &  78.42\% & 74.86\% & 98.73\% & 93.79\% & 69.07\% & 69.07\% & 90.09\% &82.98\%\\
         \cmidrule{3-10}  
          ASR-30 &  Att-2 &89.34\% &89.34\% &86.48\% &94.47\% &87.96\% &87.96\%	  &83.85\% &83.77\% \\
         \cmidrule{3-10}  
          ~ &  Att-3 &97.01\%	&94.09\% &93.90\% &95.18\% &58.71\%	&7.63\% &75.33\% &91.28\% \\
        \bottomrule [0.4mm] 
    \end{tabular}
    }
    
\end{table}
\subsection{Impact of Target Class Choice}
\label{appendixF6}

We theorized that the choice of target class might influence the attack performance of DMBA. To explore this, we conducted experiments using various target classes on CIFAR-100 and GTSRB, with the results displayed in Table \ref{Table F.1}. Interestingly, our findings indicate that the choice of target class does not significantly impact the attack performance of DMBA.The conclusion is consistent with the main text.

\renewcommand{\thetable}{F.\arabic{table}}
\setcounter{table}{0}   

\begin{table}[H]
    \renewcommand{\arraystretch}{0.5} 
    \centering 
    \tiny 
    \vspace{-2mm}
    \caption{Impact of target class choice on attack performance.}
    \label{Table F.1}
    \setlength{\tabcolsep}{1.mm}{
        \begin{tabular}{ccccc@{\hspace{3.6mm}}ccc}
         \toprule [0.4mm] 
          \multicolumn{2}{c}{Datasets} & \multicolumn{3}{c}{CIFAR-100} &  \multicolumn{3}{c}{GTSRB}\\
         \midrule  
          \multicolumn{2}{c}{Target Labels}& [15,20,25] & [4,57,89]& [26,34,90]& [0,20,29] & [3,17,25] & [10,24,40]\\
         \midrule  
         \multicolumn{2}{c}{ACC} &  52.63\% &  52.61\% &  53.64\% &  98.13\% &  97.27\% &  98.11\%\\
         \midrule  
         \multirow{3}*{~} &  Att-1 &  91.19\% &  97.32\% &  95.70\% &  98.72\% &  96.01\% &  96.09\%\\
         \cmidrule{3-8}  
          ASR &  Att-2 &95.80\%	&96.84\% &96.24\%	&99.83\%	&98.54\%	& 99.86\%	\\
         \cmidrule{3-8}  
          ~ &  Att-3 &  92.25\% &  93.21\% &  93.37\% &  99.31\% &  96.44\% & 99.88\% \\
         \midrule  
         \multirow{3}*{~} &  Att-1 & 69.07\%  & 84.89\% & 77.93\% & 96.79\% &89.28\% &	94.33\% \\
         \cmidrule{3-8}  
          ASR-30 &  Att-2 &87.96\% &72.14\% &86.28\%	 
                    &99.67\% &97.73\% &83.55\% \\
         \cmidrule{3-8}  
          ~ &  Att-3 &58.71\% &73.00\%	&73.00\% &99.25\% &92.91\% &94.30\%\\
        \bottomrule [0.4mm] 
    \end{tabular}
    }
    
\end{table}
\section{Robustness Validation on Each Dataset}

 
%

\bibliographystyle{IEEEtran}
\bibliography{sn-bibliography}

@article{1,
  title={Federated learning: Strategies for improving communication efficiency},
  author={Konecn{\`y}, Jakub and McMahan, H Brendan and Yu, Felix X and Richt{\'a}rik, Peter and Suresh, Ananda Theertha and Bacon, Dave},
  journal={arXiv preprint arXiv:1610.05492},
  volume={8},
  year={2016}
}

@inproceedings{2,
  title={Communication-efficient learning of deep networks from decentralized data},
  author={McMahan, Brendan and Moore, Eider and Ramage, Daniel and Hampson, Seth and y Arcas, Blaise Aguera},
  booktitle={Artificial intelligence and statistics},
  pages={1273--1282},
  year={2017},
  organization={PMLR}
}

@article{3,
  title={Federated machine learning: Concept and applications},
  author={Yang, Qiang and Liu, Yang and Chen, Tianjian and Tong, Yongxin},
  journal={ACM Transactions on Intelligent Systems and Technology (TIST)},
  volume={10},
  number={2},
  pages={1--19},
  year={2019},
  publisher={ACM New York, NY, USA}
}

@article{4,
  title={Efficient and secure federated learning for financial applications},
  author={Liu, Tao and Wang, Zhi and He, Hui and Shi, Wei and Lin, Liangliang and An, Ran and Li, Chenhao},
  journal={Applied Sciences},
  volume={13},
  number={10},
  pages={5877},
  year={2023},
  publisher={MDPI}
}

@article{5,
  title={Federated learning in medicine: facilitating multi-institutional collaborations without sharing patient data},
  author={Sheller, Micah J and Edwards, Brandon and Reina, G Anthony and Martin, Jason and Pati, Sarthak and Kotrotsou, Aikaterini and Milchenko, Mikhail and Xu, Weilin and Marcus, Daniel and Colen, Rivka R and others},
  journal={Scientific reports},
  volume={10},
  number={1},
  pages={12598},
  year={2020},
  publisher={Nature Publishing Group UK London}
}

@inproceedings{7,
  title={Dba: Distributed backdoor attacks against federated learning},
  author={Xie, Chulin and Huang, Keli and Chen, Pin-Yu and Li, Bo},
  booktitle={International conference on learning representations},
  year={2019}
}

@article{10,
  title={A little is enough: Circumventing defenses for distributed learning},
  author={Baruch, Gilad and Baruch, Moran and Goldberg, Yoav},
  journal={Advances in Neural Information Processing Systems},
  volume={32},
  year={2019}
}

@article{17,
  title={Multi-targeted backdoor: Indentifying backdoor attack for multiple deep neural networks},
  author={Kwon, Hyun and Yoon, Hyunsoo and Park, Ki-Woong},
  journal={IEICE TRANSACTIONS on Information and Systems},
  volume={103},
  number={4},
  pages={883--887},
  year={2020},
  publisher={The Institute of Electronics, Information and Communication Engineers}
}

@inproceedings{20,
  title={A new backdoor attack in cnns by training set corruption without label poisoning},
  author={Barni, Mauro and Kallas, Kassem and Tondi, Benedetta},
  booktitle={2019 IEEE International Conference on Image Processing (ICIP)},
  pages={101--105},
  year={2019},
  organization={IEEE}
}

@article{24,
  title={An experimental study of byzantine-robust aggregation schemes in federated learning},
  author={Li, Shenghui and Ngai, Edith C-H and Voigt, Thiemo},
  journal={IEEE Transactions on Big Data},
  year={2023},
  publisher={IEEE}
}

@article{34,
  title={Image-based crop disease detection with federated learning},
  author={Mamba Kabala, Denis and Hafiane, Adel and Bobelin, Laurent and Canals, Rapha{\"e}l},
  journal={Scientific Reports},
  volume={13},
  number={1},
  pages={19220},
  year={2023},
  publisher={Nature Publishing Group UK London}
}

@inproceedings{57,
  title={The unreasonable effectiveness of deep features as a perceptual metric},
  author={Zhang, Richard and Isola, Phillip and Efros, Alexei A and Shechtman, Eli and Wang, Oliver},
  booktitle={Proceedings of the IEEE conference on computer vision and pattern recognition},
  pages={586--595},
  year={2018}
}

@article{konevcny2016federated,
  title={Federated learning: Strategies for improving communication efficiency},
  author={Kone{\v{c}}n{\`y}, Jakub and McMahan, H Brendan and Yu, Felix X and Richt{\'a}rik, Peter and Suresh, Ananda Theertha and Bacon, Dave},
  journal={arXiv preprint arXiv:1610.05492},
  year={2016}
}

@inproceedings{mcmahan2017communication,
  title={Communication-efficient learning of deep networks from decentralized data},
  author={McMahan, Brendan and Moore, Eider and Ramage, Daniel and Hampson, Seth and y Arcas, Blaise Aguera},
  booktitle={Artificial intelligence and statistics},
  pages={1273--1282},
  year={2017},
  organization={PMLR}
}

@article{yang2019federated,
  title={Federated machine learning: Concept and applications},
  author={Yang, Qiang and Liu, Yang and Chen, Tianjian and Tong, Yongxin},
  journal={ACM Transactions on Intelligent Systems and Technology (TIST)},
  volume={10},
  number={2},
  pages={1--19},
  year={2019},
  publisher={ACM New York, NY, USA}
}

@article{imteaj2021survey,
  title={A survey on federated learning for resource-constrained IoT devices},
  author={Imteaj, Ahmed and Thakker, Urmish and Wang, Shiqiang and Li, Jian and Amini, M Hadi},
  journal={IEEE Internet of Things Journal},
  volume={9},
  number={1},
  pages={1--24},
  year={2021},
  publisher={IEEE}
}

@article{wang2023blockchain,
  title={Blockchain-empowered federated learning through model and feature calibration},
  author={Wang, Qianlong and Liao, Weixian and Guo, Yifan and Mcguire, Michael and Yu, Wei},
  journal={IEEE Internet of Things Journal},
  volume={11},
  number={4},
  pages={5770--5780},
  year={2023},
  publisher={IEEE}
}

@article{sheller2020federated,
  title={Federated learning in medicine: facilitating multi-institutional collaborations without sharing patient data},
  author={Sheller, Micah J and Edwards, Brandon and Reina, G Anthony and Martin, Jason and Pati, Sarthak and Kotrotsou, Aikaterini and Milchenko, Mikhail and Xu, Weilin and Marcus, Daniel and Colen, Rivka R and others},
  journal={Scientific reports},
  volume={10},
  number={1},
  pages={12598},
  year={2020},
  publisher={Nature Publishing Group UK London}
}

@inproceedings{chellapandi2023survey,
  title={A survey of federated learning for connected and automated vehicles},
  author={Chellapandi, Vishnu Pandi and Yuan, Liangqi and {\.Z}ak, Stanislaw H and Wang, Ziran},
  booktitle={2023 IEEE 26th International Conference on Intelligent Transportation Systems (ITSC)},
  pages={2485--2492},
  year={2023},
  organization={IEEE}
}

@inproceedings{zhang2022neurotoxin,
  title={Neurotoxin: Durable backdoors in federated learning},
  author={Zhang, Zhengming and Panda, Ashwinee and Song, Linyue and Yang, Yaoqing and Mahoney, Michael and Mittal, Prateek and Kannan, Ramchandran and Gonzalez, Joseph},
  booktitle={International Conference on Machine Learning},
  pages={26429--26446},
  year={2022},
  organization={PMLR}
}

@article{wang2020attack,
  title={Attack of the tails: Yes, you really can backdoor federated learning},
  author={Wang, Hongyi and Sreenivasan, Kartik and Rajput, Shashank and Vishwakarma, Harit and Agarwal, Saurabh and Sohn, Jy-yong and Lee, Kangwook and Papailiopoulos, Dimitris},
  journal={Advances in neural information processing systems},
  volume={33},
  pages={16070--16084},
  year={2020}
}

@article{baruch2019little,
  title={A little is enough: Circumventing defenses for distributed learning},
  author={Baruch, Gilad and Baruch, Moran and Goldberg, Yoav},
  journal={Advances in Neural Information Processing Systems},
  volume={32},
  year={2019}
}

@article{gu2017badnets,
  title={Badnets: Identifying vulnerabilities in the machine learning model supply chain},
  author={Gu, Tianyu and Dolan-Gavitt, Brendan and Garg, Siddharth},
  journal={arXiv preprint arXiv:1708.06733},
  year={2017}
}

@article{shafahi2018poison,
  title={Poison frogs! targeted clean-label poisoning attacks on neural networks},
  author={Shafahi, Ali and Huang, W Ronny and Najibi, Mahyar and Suciu, Octavian and Studer, Christoph and Dumitras, Tudor and Goldstein, Tom},
  journal={Advances in neural information processing systems},
  volume={31},
  year={2018}
}

@inproceedings{salem2022dynamic,
  title={Dynamic backdoor attacks against machine learning models},
  author={Salem, Ahmed and Wen, Rui and Backes, Michael and Ma, Shiqing and Zhang, Yang},
  booktitle={2022 IEEE 7th European Symposium on Security and Privacy (EuroS\&P)},
  pages={703--718},
  year={2022},
  organization={IEEE}
}

@article{chen2017targeted,
  title={Targeted backdoor attacks on deep learning systems using data poisoning},
  author={Chen, Xinyun and Liu, Chang and Li, Bo and Lu, Kimberly and Song, Dawn},
  journal={arXiv preprint arXiv:1712.05526},
  year={2017}
}

@article{kwon2022multi,
  title={Multi-model selective backdoor attack with different trigger positions},
  author={Kwon, Hyun},
  journal={IEICE TRANSACTIONS on Information and Systems},
  volume={105},
  number={1},
  pages={170--174},
  year={2022},
  publisher={The Institute of Electronics, Information and Communication Engineers}
}

@article{xue2022imperceptible,
  title={Imperceptible and multi-channel backdoor attack against deep neural networks},
  author={Xue, Mingfu and Ni, Shifeng and Wu, Yinghao and Zhang, Yushu and Wang, Jian and Liu, Weiqiang},
  journal={arXiv preprint arXiv:2201.13164},
  year={2022}
}

@article{hou2022m,
  title={M-to-n backdoor paradigm: A stealthy and fuzzy attack to deep learning models},
  author={Hou, Linshan and Hua, Zhongyun and Li, Yuhong and Zhang, Leo Yu},
  journal={arXiv preprint arXiv:2211.01875},
  year={2022}
}

@inproceedings{bagdasaryan2020backdoor,
  title={How to backdoor federated learning},
  author={Bagdasaryan, Eugene and Veit, Andreas and Hua, Yiqing and Estrin, Deborah and Shmatikov, Vitaly},
  booktitle={International conference on artificial intelligence and statistics},
  pages={2938--2948},
  year={2020},
  organization={PMLR}
}

@article{kwon2020multi,
  title={Multi-targeted backdoor: Indentifying backdoor attack for multiple deep neural networks},
  author={Kwon, Hyun and Yoon, Hyunsoo and Park, Ki-Woong},
  journal={IEICE TRANSACTIONS on Information and Systems},
  volume={103},
  number={4},
  pages={883--887},
  year={2020},
  publisher={The Institute of Electronics, Information and Communication Engineers}
}

@inproceedings{barni2019new,
  title={A new backdoor attack in cnns by training set corruption without label poisoning},
  author={Barni, Mauro and Kallas, Kassem and Tondi, Benedetta},
  booktitle={2019 IEEE International Conference on Image Processing (ICIP)},
  pages={101--105},
  year={2019},
  organization={IEEE}
}

@article{xue2020one,
  title={One-to-N \& N-to-One: Two advanced backdoor attacks against deep learning models},
  author={Xue, Mingfu and He, Can and Wang, Jian and Liu, Weiqiang},
  journal={IEEE Transactions on Dependable and Secure Computing},
  volume={19},
  number={3},
  pages={1562--1578},
  year={2020},
  publisher={IEEE}
}

@inproceedings{bhagoji2019analyzing,
  title={Analyzing federated learning through an adversarial lens},
  author={Bhagoji, Arjun Nitin and Chakraborty, Supriyo and Mittal, Prateek and Calo, Seraphin},
  booktitle={International conference on machine learning},
  pages={634--643},
  year={2019},
  organization={PMLR}
}

@inproceedings{xie2019dba,
  title={Dba: Distributed backdoor attacks against federated learning},
  author={Xie, Chulin and Huang, Keli and Chen, Pin-Yu and Li, Bo},
  booktitle={International conference on learning representations},
  year={2019}
}

@inproceedings{liu2024beyond,
  title={Beyond traditional threats: A persistent backdoor attack on federated learning},
  author={Liu, Tao and Zhang, Yuhang and Feng, Zhu and Yang, Zhiqin and Xu, Chen and Man, Dapeng and Yang, Wu},
  booktitle={Proceedings of the AAAI Conference on Artificial Intelligence},
  volume={38},
  number={19},
  pages={21359--21367},
  year={2024}
}

@article{wang2024dual,
  title={Dual model replacement: invisible multi-target backdoor attack based on federal learning},
  author={Wang, Rong and Zhou, Guichen and Gao, Mingjun and Xiao, Yunpeng},
  journal={arXiv preprint arXiv:2404.13946},
  year={2024}
}

@article{fung1808mitigating,
  author       = {Clement Fung and
                  Chris J. M. Yoon and
                  Ivan Beschastnikh},
  title        = {Mitigating Sybils in Federated Learning Poisoning},
  journal      = {CoRR},
  volume       = {abs/1808.04866},
  year         = {2018},
}

@inproceedings{zhang2022fldetector,
  title={Fldetector: Defending federated learning against model poisoning attacks via detecting malicious clients},
  author={Zhang, Zaixi and Cao, Xiaoyu and Jia, Jinyuan and Gong, Neil Zhenqiang},
  booktitle={Proceedings of the 28th ACM SIGKDD conference on knowledge discovery and data mining},
  pages={2545--2555},
  year={2022}
}

@inproceedings{xu2022byzantine,
  title={Byzantine-robust federated learning through collaborative malicious gradient filtering},
  author={Xu, Jian and Huang, Shao-Lun and Song, Linqi and Lan, Tian},
  booktitle={2022 IEEE 42nd International Conference on Distributed Computing Systems (ICDCS)},
  pages={1223--1235},
  year={2022},
  organization={IEEE}
}

@article{geyer2017differentially,
  title={Differentially private federated learning: A client level perspective},
  author={Geyer, Robin C and Klein, Tassilo and Nabi, Moin},
  journal={arXiv preprint arXiv:1712.07557},
  year={2017}
}

@article{mcmahan2017learning,
  title={Learning differentially private recurrent language models},
  author={McMahan, H Brendan and Ramage, Daniel and Talwar, Kunal and Zhang, Li},
  journal={arXiv preprint arXiv:1710.06963},
  year={2017}
}

@article{li2023experimental,
  title={An experimental study of byzantine-robust aggregation schemes in federated learning},
  author={Li, Shenghui and Ngai, Edith C-H and Voigt, Thiemo},
  journal={IEEE Transactions on Big Data},
  year={2023},
  publisher={IEEE}
}

@article{wu2020mitigating,
  title={Mitigating backdoor attacks in federated learning},
  author={Wu, Chen and Yang, Xian and Zhu, Sencun and Mitra, Prasenjit},
  journal={arXiv preprint arXiv:2011.01767},
  year={2020}
}

@article{wu2022federated,
  title={Federated unlearning with knowledge distillation},
  author={Wu, Chen and Zhu, Sencun and Mitra, Prasenjit},
  journal={arXiv preprint arXiv:2201.09441},
  year={2022}
}

@article{mamba2023image,
  title={Image-based crop disease detection with federated learning},
  author={Mamba Kabala, Denis and Hafiane, Adel and Bobelin, Laurent and Canals, Rapha{\"e}l},
  journal={Scientific Reports},
  volume={13},
  number={1},
  pages={19220},
  year={2023},
  publisher={Nature Publishing Group UK London}
}

@inproceedings{tolpegin2020data,
  title={Data poisoning attacks against federated learning systems},
  author={Tolpegin, Vale and Truex, Stacey and Gursoy, Mehmet Emre and Liu, Ling},
  booktitle={Computer security--ESORICs 2020: 25th European symposium on research in computer security, ESORICs 2020, guildford, UK, September 14--18, 2020, proceedings, part i 25},
  pages={480--501},
  year={2020},
  organization={Springer}
}

@article{shannon1949communication,
  title={Communication theory of secrecy systems},
  author={Shannon, Claude E},
  journal={The Bell system technical journal},
  volume={28},
  number={4},
  pages={656--715},
  year={1949},
  publisher={Nokia Bell Labs}
}

@inproceedings{wang2022invisible,
  title={An invisible black-box backdoor attack through frequency domain},
  author={Wang, Tong and Yao, Yuan and Xu, Feng and An, Shengwei and Tong, Hanghang and Wang, Ting},
  booktitle={European Conference on Computer Vision},
  pages={396--413},
  year={2022},
  organization={Springer}
}

@inproceedings{yu2023backdoor,
  title={Backdoor attacks against deep image compression via adaptive frequency trigger},
  author={Yu, Yi and Wang, Yufei and Yang, Wenhan and Lu, Shijian and Tan, Yap-Peng and Kot, Alex C},
  booktitle={Proceedings of the IEEE/CVF Conference on Computer Vision and Pattern Recognition},
  pages={12250--12259},
  year={2023}
}

@inproceedings{feng2022fiba,
  title={Fiba: Frequency-injection based backdoor attack in medical image analysis},
  author={Feng, Yu and Ma, Benteng and Zhang, Jing and Zhao, Shanshan and Xia, Yong and Tao, Dacheng},
  booktitle={Proceedings of the IEEE/CVF Conference on Computer Vision and Pattern Recognition},
  pages={20876--20885},
  year={2022}
}

@inproceedings{zeng2021rethinking,
  title={Rethinking the backdoor attacks' triggers: A frequency perspective},
  author={Zeng, Yi and Park, Won and Mao, Z Morley and Jia, Ruoxi},
  booktitle={Proceedings of the IEEE/CVF international conference on computer vision},
  pages={16473--16481},
  year={2021}
}

@article{hou2023stealthy,
  title={A stealthy and robust backdoor attack via frequency domain transform},
  author={Hou, Ruitao and Huang, Teng and Yan, Hongyang and Ke, Lishan and Tang, Weixuan},
  journal={World Wide Web},
  volume={26},
  number={5},
  pages={2767--2783},
  year={2023},
  publisher={Springer}
}

@inproceedings{xiao2022invisible,
  title={An Invisible Backdoor Attack based on DCT-Injection},
  author={Xiao, Tao and Deng, Xinyang and Jiang, Wen},
  booktitle={2022 IEEE International Conference on Unmanned Systems (ICUS)},
  pages={399--404},
  year={2022},
  organization={IEEE}
}

@article{lesort2020continual,
  title={Continual learning: Tackling catastrophic forgetting in deep neural networks with replay processes},
  author={Lesort, Timoth{\'e}e},
  journal={arXiv preprint arXiv:2007.00487},
  year={2020}
}

@inproceedings{liu2019end,
  title={End-to-end multi-task learning with attention},
  author={Liu, Shikun and Johns, Edward and Davison, Andrew J},
  booktitle={Proceedings of the IEEE/CVF conference on computer vision and pattern recognition},
  pages={1871--1880},
  year={2019}
}

@inproceedings{hessel2018rainbow,
  title={Rainbow: Combining improvements in deep reinforcement learning},
  author={Hessel, Matteo and Modayil, Joseph and Van Hasselt, Hado and Schaul, Tom and Ostrovski, Georg and Dabney, Will and Horgan, Dan and Piot, Bilal and Azar, Mohammad and Silver, David},
  booktitle={Proceedings of the AAAI conference on artificial intelligence},
  volume={32},
  number={1},
  year={2018}
}

@article{paszke1912pytorch,
  title={Pytorch: An imperative style, high-performance deep learning library. arXiv 2019},
  author={Paszke, Adam and Gross, Sam and Massa, Francisco and Lerer, Adam and Bradbury, James and Chanan, Gregory and Killeen, Trevor and Lin, Zeming and Gimelshein, Natalia and Antiga, Luca and others},
  journal={arXiv preprint arXiv:1912.01703},
  volume={10},
  year={1912}
}

@article{krizhevsky2009learning,
  title={Learning multiple layers of features from tiny images},
  author={Krizhevsky, Alex and Hinton, Geoffrey and others},
  year={2009},
  publisher={Toronto, ON, Canada}
}

@inproceedings{stallkamp2011german,
  title={The German traffic sign recognition benchmark: a multi-class classification competition},
  author={Stallkamp, Johannes and Schlipsing, Marc and Salmen, Jan and Igel, Christian},
  booktitle={The 2011 international joint conference on neural networks},
  pages={1453--1460},
  year={2011},
  organization={IEEE}
}

@inproceedings{he2016deep,
  title={Deep residual learning for image recognition},
  author={He, Kaiming and Zhang, Xiangyu and Ren, Shaoqing and Sun, Jian},
  booktitle={Proceedings of the IEEE conference on computer vision and pattern recognition},
  pages={770--778},
  year={2016}
}

@article{khayam2003discrete,
  title={The discrete cosine transform (DCT): theory and application},
  author={Khayam, Syed Ali},
  journal={Michigan State University},
  volume={114},
  number={1},
  pages={31},
  year={2003}
}

@inproceedings{li2023poisoning,
  title={Poisoning-based backdoor attacks in computer vision},
  author={Li, Yiming},
  booktitle={Proceedings of the AAAI Conference on Artificial Intelligence},
  volume={37},
  number={13},
  pages={16121--16122},
  year={2023}
}

@inproceedings{jiang2023color,
  title={Color backdoor: A robust poisoning attack in color space},
  author={Jiang, Wenbo and Li, Hongwei and Xu, Guowen and Zhang, Tianwei},
  booktitle={Proceedings of the IEEE/CVF conference on computer vision and pattern recognition},
  pages={8133--8142},
  year={2023}
}

@article{wang2004image,
  title={Image quality assessment: from error visibility to structural similarity},
  author={Wang, Zhou and Bovik, Alan C and Sheikh, Hamid R and Simoncelli, Eero P},
  journal={IEEE transactions on image processing},
  volume={13},
  number={4},
  pages={600--612},
  year={2004},
  publisher={IEEE}
}

@inproceedings{zhang2018unreasonable,
  title={The unreasonable effectiveness of deep features as a perceptual metric},
  author={Zhang, Richard and Isola, Phillip and Efros, Alexei A and Shechtman, Eli and Wang, Oliver},
  booktitle={Proceedings of the IEEE conference on computer vision and pattern recognition},
  pages={586--595},
  year={2018}
}

@article{wang2022dispersed,
  title={Dispersed pixel perturbation-based imperceptible backdoor trigger for image classifier models},
  author={Wang, Yulong and Zhao, Minghui and Li, Shenghong and Yuan, Xin and Ni, Wei},
  journal={IEEE Transactions on Information Forensics and Security},
  volume={17},
  pages={3091--3106},
  year={2022},
  publisher={IEEE}
}

@inproceedings{kwon2020targetnet,
  title={TargetNet backdoor: attack on deep neural network with use of different triggers},
  author={Kwon, Hyun and Roh, Jungmin and Yoon, Hyunsoo and Park, Ki-Woong},
  booktitle={Proceedings of the 2020 5th International Conference on Intelligent Information Technology},
  pages={140--145},
  year={2020}
}

@article{lu2024towards,
  title={Towards Secure Internet of Things-Enabled Intelligent Transportation Systems: A Comprehensive Review.},
  author={Lu, Changxia and Wang, Fengyun},
  journal={International Journal of Advanced Computer Science \& Applications},
  volume={15},
  number={7},
  year={2024}
}

@article{ryan2025smart,
  title={Smart Surveillance: Identifying IoT Device Behaviours using ML-Powered Traffic Analysis},
  author={Ryan, Reza and Paciente, Napoleon and Youngs, Cahil and Karie, Nickson and Li, Qian and Ferdosian, Nasim},
  journal={arXiv preprint arXiv:2512.13709},
  year={2025}
}

@article{zhukabayeva2025cybersecurity,
  title={Cybersecurity solutions for industrial internet of things--edge computing integration: Challenges, threats, and future directions},
  author={Zhukabayeva, Tamara and Zholshiyeva, Lazzat and Karabayev, Nurdaulet and Khan, Shafiullah and Alnazzawi, Noha},
  journal={Sensors},
  volume={25},
  number={1},
  pages={213},
  year={2025},
  publisher={MDPI}
}

@misc{kolias2017ddos,
  title={DDoS in the IoT: Mirai and other botnets. Computer (Long Beach Calif) 50: 80--84},
  author={Kolias, C and Kambourakis, G and Stavrou, A and Voas, J},
  year={2017}
}

@inproceedings{nguyen2019diot,
  title={D{\"I}oT: A federated self-learning anomaly detection system for IoT},
  author={Nguyen, Thien Duc and Marchal, Samuel and Miettinen, Markus and Fereidooni, Hossein and Asokan, Nadarajah and Sadeghi, Ahmad-Reza},
  booktitle={2019 IEEE 39th International conference on distributed computing systems (ICDCS)},
  pages={756--767},
  year={2019},
  organization={IEEE}
}

@article{ladisa2022taxonomy,
  title={Taxonomy of Attacks on Open-Source Software Supply Chains, 2022},
  author={Ladisa, Piergiorgio and Plate, Henrik and Martinez, Matias and Barais, Olivier},
  journal={URL http://arxiv. org/abs/2204.04008},
  year={2022}
}

@article{gong2022coordinated,
  title={Coordinated backdoor attacks against federated learning with model-dependent triggers},
  author={Gong, Xueluan and Chen, Yanjiao and Huang, Huayang and Liao, Yuqing and Wang, Shuai and Wang, Qian},
  journal={IEEE network},
  volume={36},
  number={1},
  pages={84--90},
  year={2022},
  publisher={IEEE}
}

@article{alharbi2024collusive,
  title={Collusive backdoor attacks in federated learning frameworks for IoT systems},
  author={Alharbi, Saier and Guo, Yifan and Yu, Wei},
  journal={IEEE Internet of Things Journal},
  volume={11},
  number={11},
  pages={19694--19707},
  year={2024},
  publisher={IEEE}
}

@article{pan2024one,
  title={One-shot backdoor removal for federated learning},
  author={Pan, Zijie and Ying, Zuobin and Wang, Yajie and Zhang, Chuan and Li, Chunhai and Zhu, Liehuang},
  journal={IEEE Internet of Things Journal},
  volume={11},
  number={23},
  pages={37718--37730},
  year={2024},
  publisher={IEEE}
}

@inproceedings{bonazzi2025picosam2,
  title={PicoSAM2: Low-Latency Segmentation In-Sensor for Edge Vision Applications},
  author={Bonazzi, Pietro and Farronato, Nicola and Zihlmann, Stefan and Qin, Haotong and Magno, Michele},
  booktitle={2025 IEEE SENSORS},
  pages={1--4},
  year={2025},
  organization={IEEE}
}

@inproceedings{dutto2024collaborative,
  title={Collaborative visual place recognition through federated learning},
  author={Dutto, Mattia and Berton, Gabriele and Caldarola, Debora and Fan{\`\i}, Eros and Trivigno, Gabriele and Masone, Carlo},
  booktitle={Proceedings of the IEEE/CVF Conference on Computer Vision and Pattern Recognition},
  pages={4215--4225},
  year={2024}
}

@article{wang2024defending,
  title={Defending against data and model backdoor attacks in federated learning},
  author={Wang, Hao and Mu, Xuejiao and Wang, Dong and Xu, Qiang and Li, Kaiju},
  journal={IEEE Internet of Things Journal},
  volume={11},
  number={24},
  pages={39276--39294},
  year={2024},
  publisher={IEEE}
}

@article{n2025real,
  title={Real-time traffic sign recognition and autonomous vehicle control system using convolutional neural networks},
  author={N. G, Girish Kumar and Kishore, Ashish and Krishna, Aaditya J},
  journal={Multimedia Tools and Applications},
  volume={84},
  number={34},
  pages={43119--43154},
  year={2025},
  publisher={Springer}
}


 





\end{document}